\documentclass[letterpaper]{article} 
\usepackage{aaai24}  
\usepackage{times}  
\usepackage{helvet}  
\usepackage{courier}  
\usepackage[hyphens]{url}  
\usepackage{graphicx} 
\usepackage{natbib}
\usepackage{caption} 
\usepackage{todonotes}
\usepackage{algorithm}
\usepackage[noend]{algorithmic}
\usepackage{enumitem}
\usepackage{subfig}
\usepackage{wrapfig}

\usepackage{newfloat}
\usepackage{listings}
\DeclareCaptionStyle{ruled}{labelfont=normalfont,labelsep=colon,strut=off} 
\floatstyle{ruled}
\newfloat{listing}{tb}{lst}{}
\floatname{listing}{Listing}
\usepackage{subdef}

\usepackage{xspace}

\title{Generator Assisted  Mixture of Experts  For Feature Acquisition in Batch}
\author{
    Vedang Asgaonkar, Aditya Jain, Abir De
}
\affiliations{IIT Bombay\\\{vedang, adityajainjhs, abir\}@cse.iitb.ac.in}
 \makeatletter
\g@addto@macro{\normalsize}{%
\setlength{\abovedisplayskip}{1.5pt plus1pt}%
\setlength{\abovedisplayshortskip}{1.5pt plus1pt}%
\setlength{\belowdisplayskip}{1.5pt plus1pt}%
\setlength{\belowdisplayshortskip}{1.5pt plus1pt}}
\let\c@table\c@figure
\makeatother
\renewcommand{\cite}{\citep}
\usepackage[capitalize,noabbrev]{cleveref}

\begin{document}

\maketitle

\begin{abstract}
Given a set of observations, feature acquisition is about finding the subset of unobserved features which would enhance accuracy. Such problems have been explored in a sequential setting in prior work. Here, the model receives feedback from every new feature acquired and chooses to explore more features or to predict. However, sequential acquisition is not feasible in some settings where time is of the essence. We consider the problem of feature acquisition in batch, where the subset of features to be queried in batch is chosen based on the currently observed features, and then acquired as a batch, followed by prediction. We solve this problem using several technical innovations. First, we use a feature generator to draw a subset of the synthetic features for some examples, which reduces the cost of oracle queries. Second, to make the feature acquisition problem tractable for the large heterogeneous observed features, we partition the data into buckets, by borrowing tools from locality sensitive hashing and then train a mixture of experts model. Third, we design a tractable lower bound of the original objective.
We use a greedy algorithm combined with model training to solve the underlying problem.
Experiments with four datasets show that our approach outperforms these methods in terms of trade-off between accuracy and feature acquisition cost. 
% Our model is able to provide significant cost saving without loss of accuracy as demonstrated in our ablation study.
\end{abstract}

\vspace{-3mm}
 \section{Introduction}
% \vspace{-1mm}

Supervised learning algorithms often assume access to a complete set of features $\xb \in \RR^d$
to model the underlying classifier $\Pr(y\given \xb)$.
However, in applications like healthcare, information retrieval, \etc, a key goal is feature acquisition~\cite{babu2016wrapper, geng2007feature}, where the learner may observe only a subset of features $\obs \subset \set{1,..,d}$ and the goal is to query for a new subset $\unobs$ from the unobserved set of features: $\unobs \subset \set{1,...,d} \cp \obs$. For example, when a patient visits a doctor with a new health issue, the doctor can observe only few symptoms. If the symptoms are not informative enough to diagnose a disease with high confidence, the doctor may ask for additional medical tests.
% In information retrieval, the search system can ask questions to the user, so that it can return more relevant content~\cite{videoapp}.

% \vspace{-1mm}
\subsection{Prior work and their limitations}
Driven by these motivations, feature acquisition is widely studied in literature. Earlier works  used tools from active learning techniques~\cite{act1,act2,act3,gong2019icebreaker,eddi}, which optimize measures based on variance, uncertainty or information gain. To improve their performance, a recent line of work   explicitly optimizes the prediction accuracy~\cite{jafa,acflow,gsm,jaaai,jml,dulac2012sequential,l1,l2,hu2018survey,yu2016scalable}, predominantly using reinforcement learning (RL).

The   above methods are tailored for \emph{sequential} feature acquisition. In such scenarios, it is feasible to observe the value of a newly acquired feature immediately after its acquisition, allowing the use of its true value to inform the acquisition of additional features. However, certain situations involve a substantial delay between querying one feature and observing its value. In these cases, it may be more practical to batch-query a subset of features instead of acquiring them one by one in an online fashion. For instance, in healthcare, the analysis of pathological samples can introduce significant delays after collection. Thus, doctors may need to obtain results from multiple tests at once for rapid diagnosis.
We provide a detailed survey of related work in Appendix~\ref{app:related}.
\subsection{Our contributions}
% \vspace{-1mm}

Responding to the above challenge,   we propose \our, a novel feature acquisition method to acquire features in batch. 
% generator assisted mixture of experts model for
% acquisition.
Specifically, we make the following contributions.

\xhdr{Using feature generator to reduce oracle queries}
% Feature generators are prevalent in the feature acquisition task. They are used to guide the feature acquisition algorithms, \eg, the feature selection policy~\cite{gsm,eddi}. 
% However, the generated features have not been used for the final label prediction. In this work, instead of querying all the features from the oracle, we draw a part of features the from the generator and directly use them for classification.
% This reduces the number of oracle queries, while giving marginal loss in accuracy. 
Feature generators are commonly used in feature acquisition tasks to guide feature selection policies~\cite{gsm,eddi}. However, these generated features typically are not utilized for final label prediction. In our work, instead of querying all features from an oracle, we draw a feature subset from the generator and directly employ them for classification, reducing the number of oracle queries with only a marginal loss in accuracy.

\xhdr{Mixture of experts on heterogeneous feature space} 
The observed features $\obs$ can vary significantly across instances.  This leads to a diverse set of acquired features and, consequently, a range of heterogeneous data domains. Generalizing across such heterogeneity using a single model is challenging. 
To address this, we partition the dataset into clusters or domains using a random hyperplane-based approximate nearest neighbor technique~\cite{indyk1999approximate}. We then build a mixture of experts model on these clusters, with each cluster representing instances likely to share a similar set of optimal features for acquisition. Each mixture component specializes in generalizing on a specific data subset.

% In principle, the observed features $\obs$ may vary widely across instances. This leads to a diverse set of acquired features, which, in turn, results in a number of heterogeneous domains in the data.  As a result, it becomes difficult to generalize across such heterogeneous data using one single model. To overcome this the bottleneck, we first partition the dataset into different clusters or domains, using random hyperplane based approximate nearest neighbor technique~\cite{indyk1999approximate} and build a mixture of expert models on these clusters. Here, each cluster consists of instances which are likely to admit the similar set of optimal features to be acquired. Each mixture component is tailored to generalize on a specific portion of the data. 

\xhdr{Discrete continuous training framework}
The original feature acquisition problem is intractable due to the coupling of a large number of set variables. To tackle this, we design a surrogate loss guided by generator confidence and seek to minimize it alongside the feature subsets. This leads to a discrete-continuous optimization framework which is NP-hard. To tackle this problem, we reformulate it into a cardinality-constrained set function optimization task. Subsequently, we  introduce novel set function properties, $(\mmin,\mmax)$-partial monotonicity and $(\gmin,\gmax)$-weak submodularity, extending recent notions of partial monotonicity~\cite{mualem2022using} and  approximate submodularity~\cite{elenberg2018restricted,harshaw2019submodular,cuha}. These properties allow us to design a greedy algorithm \our, to compute near-optimal feature subsets, with new approximation guarantee.

We experiment with four real datasets and show that, \our\ outperforms several baselines. Moreover, our extensive ablation study shows that our use of a generative model reduces the cost of querying at a minimal accuracy drop. 
%

% \vspace{-1mm}
\section{Problem formulation}
% In this section we first introduce the notations and then describe a brief overview of our goal.
% \vspace{-1mm}

\xhdr{Notations and problem setup}
% We use $\xb\in \RR^n$ for a feature vector and $y\in \Ycal$ for the associated label. We denote  $\Ical = [n]:=\set{1,..,n}$ as the set of feature 
% indices. We use $\obs\subset \Ical$ to denote the indices of observed features and 
% $\unobs \subseteq \Ical\cp \obs$ as indices of the features to be queried. Given a  
% feature $\xb$, we denote $\xb[\obs]$ as the set of features indexed by the entries 
% in the set $\obs\subset \Ical$.
% We use $x[s] \in \RR$ (non-boldfaced "x") to denote the 
% singleton feature indexed by an index $s$.  
We use $\xb\in \RR^n$ to represent a feature vector and $y\in \Ycal$ for the associated label. $\Ical = [n]$ denotes the set of feature indices. $\obs \subset \Ical$ represents the indices of observed features, while $\unobs \subseteq \Ical\cp \obs$ represents indices of subset of features to be queried. Given a feature vector $\xb$, $\xb[\obs]$ consists of features indexed by $\obs$. We denote a singleton feature as $x[s] \in \RR$ for index $s$.

Our classifier is denoted as $h \in \Hcal$, where $h(\xb)[y] = \Pr(y \given \xb)$ and $\Hcal$ is the hypothesis class. We also employ a generative model for features, denoted as $p(\xb[\sub] \given \xb[\obs])$, which generates new features $\xb[\sub]$ conditioned on observed features $\xb[\obs]$. For clear disambiguation from oracle features $\xb[\bullet]$, we use $\xg[\bullet]$ for features drawn from the generator $p$. We utilize it to draw a subset of unseen features $\sub \subset \unobs$, rather than querying the oracle.
In our work, we use the cross-entropy loss $\ell(h(\xb),y)$. 
\xhdr{High level objective} 
% \vspace{-1mm}
Given an instance $\xb$, we initially observe only a subset of the features $\xb[\obs]$ indexed by  $\obs$ which varies across the instances. In general, this small subset is not sufficient for accurate prediction. Hence, we would seek to query new features $\xb[\unobs]$ subject to a maximum number of oracle queries. Thus, our key goal is to use $\xb[\obs]$ to determine the optimal choice of $\unobs$ among all such possible subsets, such that $\xb[\obs \cup \unobs]$ results in high accuracy. Note that, here, we aim to acquire the oracle features in \emph{batch} and not in sequence, \ie, we may not observe a part of the unobserved features, before we query the rest.

Now, suppose that by some means, we have determined such subset $\unobs$, so that  $\xb[\unobs]$ obtained via querying from the oracle would result in high accuracy.
Still, it may not be always necessary to query the value of every feature $x(u)$ for all $u\in \unobs$ from the oracle. For some subset $\sub \subset \unobs$,  the predicted features $\xg[\sub]$, which are drawn the feature generator $p$ can lead to similar accuracy as the oracle features $\xb[\sub]$. Now, since cost is only involved in oracle queries, generating $\xg[\sub]$ from the generator leads to a reduced cost on $\unobs\cp \sub$.
Here, we aim to draw $\xg[\sub]$ from the feature generator $p_{\phi}$, conditioned on the observed features $\xb[\obs]$ and the rest of the features $\xb[\unobs \cp \sub]$, where the latter is queried from the oracle. Formally, we have
$\xg[\sub] \sim p  (\bullet \given \xb[\obs])$.

% \vspace{-0.5mm}
\xhdr{Problem statement} 
% \vspace{-1mm}
During training, we are given the architectures of a classifier $h$ and the feature generator $p$ as well as the training set $\set{(\xb_i,y_i,\obs_i)}_{i\in D}$ and a budget $q_{\max}$ for maximum number of oracle queries for each instance. The budget is per instance since test instances occur in isolation. Our goal is to train $h$ and $p$, as well as simultaneously compute the optimal values of $\unobs_i$  and $\sub_i$ for each $i\in D $ and $|\unobs_i \cp \sub_i| \le q_{\max}$,
so that the oracle features $\xb_i[\obs_i \cup \unobs_i \cp \sub_i]$  and the generated features $\xg_i[\sub_i] \sim p(\bullet \given \xb[\obs_i ])$ provide high accuracy on $h$. In theory, one can encode the above task in the following training 
% problem:
loss:
\begin{align}
 &\loss(h,p; \unobs_i, \sub_i \given  \obs_i)\nn\\
 &= \EE_{\xg_i [\sub_i] \sim    p(\bullet \given \xb_i[\obs_i  ])} \Big[ \ell\left(h\big(\xb_i [\obs_i \cup \unobs_i \cp\sub_i] \cup \xg_i [\sub_i]\big), y_i\right)   \Big] \label{eq:loss}; 
\end{align}
and solve the following optimization problem.
\begin{align}
&     \mini_{h,p, \set{\sub_i \subseteq \Ical, \unobs_i \subseteq \Ical}_{i\in D}}\ \sum_{i\in D} \loss(h,p; \unobs_i, \sub_i \given  \obs_i)  \label{eq:obj}  \\
% L(h,p; \unobs_i, \sub_i \given  \obs_i) 
& \quad \text{subject to,} \quad   |\unobs_i \cp \sub_i| \le q_{\max} \ \ \text{for all } i\in D \label{eq:conn}
\end{align}
% \begin{align}
% &    \mini_{h,p, \set{\sub_i \subseteq \Ical, \unobs_i \subseteq \Ical}_{i\in D}}\ \sum_{i\in D} \EE_{\xg_i [\sub_i] \sim    p(\bullet \given \xb_i[\obs_i  ])} \Big[ \ell\left(h\big(\xb_i [\obs_i \cup \unobs_i \cp\sub_i] \cup \xg_i [\sub_i]\big), y_i\right)   \Big]\label{eq:obj} \\
% % L(h,p; \unobs_i, \sub_i \given  \obs_i) 
% & \qquad \text{subject to,} \   |\unobs_i \cp \sub_i| \le b \ \ \text{for all } i\in D \label{eq:con}
% \end{align}
If we could solve this problem, then, for a test instance, we can directly use the optimal $\unobs^* _i$ and $\sub^* _i$ from the nearest training example.

% Note that, we keep the same the budget for all $i\in D$ for brevity. Our method can also be applied in cases, where the budget is varied across instances $|\unobs_i \cp \sub_i| \le {q}_i$ or where the budget is imposed at the aggregated level, \ie, $\sum_{i\in D} |\unobs_i \cp \sub_i| \le q_{\max} |D|$. 
%
There is no cost or budget associated with drawing features from the generator. Therefore, in principle, $\sub_i$ can be as large as possible. However, a large $\sub_i$ may not always be an optimal choice in practice, because the generator may be inaccurate. For example, even if the generated feature $\xg[\sub]$ is close to its gold value $\xb[\sub]$, a small  difference $|\xg[\sub]-\xb[\sub]|$ may manifest in large prediction error~\cite{szegedy2013intriguing,goodfellow2014explaining}. 

% The interaction between the generator and the classifier may give rise different scenarios. (i) The generator can be imperfect meaning that the generated features that are significantly inaccurate
% We also observed that in some cases, the generated features are close to the oracle features. However, the classifier can still perform poorly in this case, because indeed such small differences can be manifest in significant classification error (Szedgy, Goodfellow) due to this adversarial perturbation.

% Spurious effect. It may be possible that the generator generates images with the correct object but with different surroundings. The classifier can mess up here.

%  imperfect. Moreover, its quality can vary across  different feature values $\xg[s]$ for $s\in \Ical$. As a result, merely drawing a large number of features $\xg[\sub]$ may not necessarily increase the final accuracy.  

\xhdr{Bottlenecks}
The above optimization problem involves simultaneous model training and selection of a large number of subsets  $\set{\unobs_i,\sub_i}$. As a result, it suffers from several bottlenecks as described below. 

\noindent\emph{--- (1) Large number of sets as optimization variables:} The observed subset $\obs_i$ varies widely across $i\in D$. Moreover,  the observed feature values $\xb_i[\obs]$ for the same $\obs$ also vary across instances. Hence, the optimal choice of $\unobs_i,\sub_i$ varies across instances, leading to $O(|D|)$ optimization variables. 

\noindent\emph{--- (2) Heterogeneous feature space:}
The final set of features that are fed into the classifier $\xb_i[\obs_i \cup \unobs_i\cp \sub_i]$ are very diverse, owing to a large variety of $\obs_i$, $\unobs_i$ and $\sub_i$. This results in a number of heterogeneous domains, which makes it difficult for one single model   to generalize across the entire data.

\noindent\emph{---(3) Coupling between different optimization variables:} 
Two types of couplings exist between optimization variables $\ucal{i}$ and $\vcal{i}$. Given one instance $i\in D$, the optimization variables $\ucal{i}$ and $\vcal{i}$ are coupled and so are $\ucal{i}, \vcal{i}$ and $\ucal{j}, \vcal{j}$ for different instances $i\in D$ and $j\in D$.
This complexity renders the joint optimization problem~\eqref{eq:obj} intractable.  
% econdly, even for a single instance $i\in D$, $\unobs_i$ and $\sub_i$ serve as optimization variables, further complicating the problem.

\section{Proposed approach}
In this section, we introduce \our, a generator-assisted mixture of expert model addressing identified challenges. We present a tractable alternative to optimization problem~\eqref{eq:obj} in three steps: (I) data partitioning, (II) designing mixture models, and (III) decoupling cross-instance coupling of optimization variables. Finally, we offer a set   function centric characterization of this alternative optimization and a greedy algorithm for its solution.

\subsection{Data partitioning}
% \vspace{-1mm}
In the first step, we reduce the number of optimization variables (bottleneck 1) and transform the heterogeneous set of instances into homogeneous clusters (bottleneck 2). 
% leading to improved generalization.  
% This allows us to reduce the number of optimization variables (bottleneck 1) and develop the mixture models, leading to improved generalization (bottleneck 2). 
% The key reason behind the large number of optimization variables is   fine grained choices of $\ucal{i}$ and $\vcal{i}$ for each instance $i\in D$. 
% Here, we coarsen the estimate of these sets, by assigning the same optimization variables $(\ucal{b},\vcal{b})$ for all the observed features falling in the same bucket $b$. This reduces the number of optimization variables to $O(B)$.

% In the first step, we reduce the number of optimization variables and transform the heterogeneous set of instances into homogeneous clusters. This allows us to reduce the number of optimization variables (bottleneck 1) and develop the mixture models, leading to improved generalization (bottleneck 2). 
Clustering methods like K-means and Gaussian mixture models maximize the \emph{average} intra-cluster similarity. We observed that (Section~\ref{sec:exp}) this has led to highly suboptimal partitioning, with fewer highly similar instances in one cluster and others with moderately high similarity in different clusters. To address this, we adopt a random hyperplane-based clustering technique, mitigating bucket imbalance and achieving more equitable cluster assignments. 
% This approach is crucial for enhancing the performance of our proposed model by promoting balanced and representative cluster formation.

 % \xhdr{Random hyperplane based clustering}
 % We aim to partition the observations $\set{\xb_i[\obs_i]}_{i\in D}$ into $B$ different clusters, called as "buckets", so that the observations across (within) buckets have low (high)  similarity.
 % Here, K-means or Gaussian mixture model based clustering could serve our purpose. However, we note that their objective maximized the average (or weighted average) of the intra-cluster mean of a similarity metric (cosine and RBF respectively).  Due to this,   they often encourage few points that are very highly similar in one cluster and points which have not-so-high (but still high) similarity across other clusters. Hence, we observed that most clusters contain few instances that are of very high similarity and one left out cluster has to accommodate instances that cannot share the other clusters.  This resulted in a significant bucket skew, and we resorted to a trainable random hyperplane based clustering.
 % This is sometimes encouraged by the sensitivity of the optimization wrt seed. Thus, the left out cluster has many points and while each of the other clusters have instances that admit high intra-cluster similarity.

\xhdr{Random hyperplane based clustering}
% We partition the observations $\set{\xb_i[\obs_i]}_{i\in D}$ into $B$ distinct clusters or "buckets" based on their similarity. 
% We use random hyperplane (RH) guided approximate nearest neighbor (ANN) clustering, by leveraging tools from locality sensitive hashing~\cite{indyk1999approximate}. 
% In RH based clustering, we are given the number of clusters or buckets as $B=2^M$. Then $M$ \emph{i.i.d.} spherically distributed normal vectors are drawn, \ie, $\Wb = [\wb_1,..,\wb_M] \in \RR^{n\times M}$, with $\wb_{m}\sim N(0,\II_{n})$ for $m\in [M]$. 
% % Each  $\wb_m$ is orthogonal to a random hyperplane passing through the origin.  
% %%
% Next, we hash each observation $\xb_i[\ocal{i}]$ to a bucket $b\in [\pm 1]^{M}$ which is represented as $b = \sgn(\Wb^\top \xb_i[\ocal{i}])$. Here, we pad $\xb_i[\ocal{i}]$ with $n-|\ocal{i}|$ zeros to make it compatible for pre-multiplying with $\Wb^{\top}$.
% If the buckets are balanced, each bucket will roughly have $|D|/2^M$ instances.
% %
% As proven in the seminal paper by~\citet[Section 3]{charikar2002similarity},
% the collision probability, \ie, the probability that two observed features will share same bucket, monotonically increases with the cosine similarity between them. 
We partition observations $\set{\xb_i[\obs_i]}_{i\in D}$ into $B$ clusters using Random Hyperplane (RH) guided Approximate Nearest Neighbor (ANN) clustering, employing locality-sensitive hashing~\cite{indyk1999approximate}. For this, we generate $M$ independent spherically distributed normal vectors $\Wb = [\wb_1,..,\wb_M] \in \RR^{n\times M}$, with $\wb_{m}\sim N(0,\II_{n})$. Each $\wb_m$ defines a random hyperplane through the origin. We hash each observation $\xb_i[\ocal{i}]$ to a bucket $b = \sgn(\Wb^\top \xb_i[\ocal{i}])$, where $\xb_i[\ocal{i}]$ is padded with zeros to match dimensionality. Each bucket is thus identified by a vector of $\pm 1$ of length $M$. This implies that $B\le 2^M$. In our experiments we observe that instances are roughly uniformly distributed among the $2^M$ buckets, hence each bucket will roughly have $|D|/2^M$ instances, with $B=2^M$. The probability that two observed features will share the same bucket, increases with their cosine similarity ~\citet[Section 3]{charikar2002similarity}.

In contrast to K-means of GMM, RH has two key advantages. (1) It doesn't maximize any aggregate objective thus the assignment of one instance $\xb$ doesn't affect the cluster assignment of another instance $\xb'$
(2) The randomized algorithm encourages cluster diversity

\xhdr{Reducing the number of optimization variables} 
The key reason behind the large number of optimization variables is   fine grained choices of $\ucal{i}$ and $\vcal{i}$ for each instance $i\in D$. 
Here, we coarsen the estimate of these sets, by assigning the same optimization variables $(\ucal{b},\vcal{b})$ for all the observed features falling in the same bucket $b$. This reduces the number of optimization variables to $O(B)$.

For a test example $\xb[\obs]$, we seek to find $\unobs_{b^*}$ and $\sub_{b^*}$, where the bucket $b^*$ was assigned to the training instance $i$ having the highest cosine similarity with $\xb[\obs]$. This bucket id can be immediately obtained by computing $b^* = \sgn(\Wb^\top \xb[\ocal{}])$, without explicit nearest neighbor search~\cite{charikar2002similarity}. 
During our experiments, we observed that the above clustering method works better than K-means or gaussian mixture clustering. Moreover, in our method, computation of $\ucal{b^*}$ and $\vcal{b^*}$ for a test instance $\xb[\ocal{}]$ admits $O(\log B )$ time complexity to compute the bucket id $b^*$, holding $n$ constant. On the other hand, K-means or gaussian clustering admits $O(B)$ complexity.

\subsection{Mixture models}
% \vspace{-1mm}
Having partitioned the data, as described above, we train a mixture of models across these clusters, where each model is tailored specifically to generalize on each cluster. This addresses bottleneck (2) and (3). 

% decouples the optimization problem~\eqref{eq:obj} into $B$ independent components.

 % Such a mixture model decouples the optimization problem~\eqref{eq:obj} into $B$ independent components. This leads us to get rid of the cross-instance coupling between the model parameters, $\ucal{}$ and $\vcal{}$ \ie, the first part of the bottleneck (3). 
\xhdr{Formulation of mixture models} Given a partitioning of $D$ into $B$ buckets, \ie,  $D = D_1 \cup D_2 \cup ... \cup D_B$, we build a mixture of $B$ independent classifiers $h_{\theta_b} $ and generators  $  p_{\phi_b}$ , parameterized with $\theta_b$ and $\phi_b$ for a bucket $b$. This reduces the joint optimization~\eqref{eq:obj} problem into the following 
\begin{align}
& \mini_{\set{\theta_b,\phi_b,\ucal{b},\vcal{b}}_{b\in [B]}} \sum_{b\in [B]}\sum_{i\in D_b}\loss(h_{\theta_b},p_{\phi_b};\ucal{b}, \vcal{b} \given \ocal{i}) \nonumber \\
& \text{such that,} \   |\ucal{b} \cp \vcal{b}| \le q_{\max}  \ \forall b\in [|B|]  \label{eq:mix}
\end{align}
% \xhdr{Enhancing generalization ability}
% Note that each $h_{\theta_b}$ and $p_{\phi_b}$ are trained on the instances on the $D_b$, independently of other buckets. Such independent training on similar instances provides us a better generalization power as compared to the training a single model on a large heterogeneous dataset. 
% This enhances the overall generalization ability of our framework.

\xhdr{Decoupling cross-instance optimization variables}
It is evident that the above optimization~\eqref{eq:mix} can be decoupled into $B$ independent components. For each bucket $b$, we minimize $\loss(h_{\theta_b},p_{\phi_b};\ucal{b} , \vcal{b} \given \ocal{i})$, separately from other buckets.  This reduces to the feature selection problem--- the goal of selecting one fixed set of features for multiple instances~\cite{elenberg2018restricted}. Thus, it leads us to overcome the cross-instance coupling between the model parameters, $\ucal{}$ and $\vcal{}$. It also facilitates distributed implementation.
\vspace{-2mm}
\subsection{Decoupling optimization tasks over $\unobs_b$ and $\sub_b$ }
% \vspace{-1mm}
\label{subsec:F}
\xhdr{Overview} $\loss(h_{\theta_b},p_{\phi_b};\ucal{b} , \vcal{b} \given \ocal{i})$--- the objective in a bucket $b$--- still involves a coupling between $\ucal{b}$, $\vcal{b}$. To overcome this, we build two optimization problems. We first compute the optimal $\unobs_b$ and then compute $\sub_b$ based on the optimal value of $\unobs_b$ obtained.  This addresses bottleneck (3).
\xhdr{Optimization over $\ucal{b}$} For the first optimization, we design a new loss function $F(\theta_b,\phi_b; \ucal{b} \given \ocal{b} )$ whose optimal value  with respect to $\ucal{b}$ for a given $\theta_b$ and $\phi_b$ is an upper bound of the corresponding model training loss of the optimization~\eqref{eq:mix}.  This loss is a combination of the prediction losses from the oracle and generated features, weighted by a prior uncertainty measure. This uncertainty is computed by pre-training the classifier and the generator on the observed data $\set{(\xb_i,y_i,\ocal{i})}_{i\in D}$.  Having computed the pre-trained classifier $h_0$ and the pre-trained generator $p_0$, we define
$\Delta_i(\ucal{b})$ as the uncertainty of the  
classifier when $h_0$ uses the generated features $\xg_i[\ucal{b}]\sim p_0(\bullet \given \ocal{i})$ for the whole set $\ucal{b}$, \ie,  $\Delta_i(\ucal{b}) = \EE_{\xg_i\sim p_{0}(\bullet \given \xb[\ocal{i}])} [1-\max_y h_{0}(\xb_i[\ocal{i}]\cup \xg_i[\ucal{b}])[y]]$. 
Thus, $\Delta_i(\ucal{b}) \in [0, 1-\frac{1}{|C|}]$. We rescale  $\Delta_i(\ucal{b})$ by dividing it with $1-\frac{1}{|C|}$, so that it lies in $[0,1]$.

 \begin{figure}[H]
\vspace{-9mm}
\begin{minipage}{0.48\textwidth}
\begin{algorithm}[H]
\small
\caption{Training}
\begin{algorithmic}[1]
    \REQUIRE Training data $\set{(\xb_i,y_i,\obs_i)}_{i\in D}$, Number of buckets $B=2^M$, the classifier architecture $h$ and generator architecture $p$.
        \STATE $\set{D_b}_{b\in[B]}\leftarrow$ \textsc{Partition}($D,B$)
        \FOR{bucket $b \in B$}
        \STATE $U_b ^*, \theta^* _b , \phi^* _b  \gets $\textsc{GreedyForU}($q_{\max},b, F, \gf)$
        \STATE $V_b ^*, \gets $\textsc{GreedyForV}($\lambda,b, \gl$)
        \ENDFOR

            \end{algorithmic}
\hrule
% \vspace{2mm}
\hrule
%%%
\begin{algorithmic}[1]
    \FUNCTION{\textsc{Partition}($D$, $B=2^M$)}
        \STATE $\Wb \leftarrow [\wb_m]_{m \in [M]}\sim N(0,\II_{n})$
        \STATE hash$[i]\leftarrow \sgn(\Wb^{\top} \xb_i [\ocal{i}])$ for all $i\in D$
        \STATE $D_b \leftarrow \set{i:\text{hash}[i]=b}$
        \STATE \textbf{Return} $\set{D_b}_{b\in [2^M]}$ 
    \ENDFUNCTION
    \end{algorithmic}

\hrule
 % %%%
\begin{algorithmic}[1]
    \FUNCTION{\textsc{GreedyForV}($\lambda,b, F, \gl$)}
    \STATE \textbf{Require:} trained models $\theta^* _b, \phi^* _b$ and the optimal subset $\ucal{b}^*$
   % \STATE  \REQUIRE Access to computation of $G_F$ and $F$
    \STATE $V_b \leftarrow \emptyset$
        % \STATE $E \leftarrow \mathcal{F}\cp \bigcup_{i \in b} \ocal{i} \cup  \ucal{b}$
         \FOR{$q\in [\lambda]$}
         \STATE $e^* \gets  {\argmin}_{e\not\in \vcal{b}\cup \ocal{i}} \ \gl(e  \given \vcal{b} ) $
          \IF{$\gl(e^*  \given \vcal{b} ) < 0$}
           \STATE $V_b \gets V_b \cup e^*$ 
           \STATE \textbf{break}
          \ENDIF
         % \STATE $V_b \gets V_b \cup e^*$ if $\gl(e  \given \vcal{b} ) < 0$ 
         %  % \STATE $\theta^* _b , \phi^* _b  \gets \textsc{Train}(F,\ucal{b},b)$
         \ENDFOR
        % \STATE $e^* \leftarrow \underset{e}{\mathrm{argmin}}\sum_{i\in b}F(h_{\theta_b},p_{\phi_b};\ucal{b}\cup\{e\} \given \ocal{i}); ~e \in E $
        \STATE \textbf{Return} $V_b$
    \ENDFUNCTION

\end{algorithmic}
\label{alg:training}
\end{algorithm}
\end{minipage} \hspace{3mm}
\end{figure}

Then, we  define the new loss $F$ as follows.
\begin{align}
& \hspace{-2mm}F(h_{\theta_b},p_{\phi_b};\ucal{b} \given \ocal{i})  =
 \Delta_i(\ucal{b})  \cdot \ell(h_{\theta_b}(\xb_i[\ocal{i} \cup \ucal{b}]), y )   \nonumber \\  
& \  + (1-\Delta_i(\ucal{b}) ) \cdot \EE_{\xg[\ucal{}] }   \ell\left(h_{\theta_b}\big(\xb_i [\obs_i ] \cup \xg_i [\ucal{b}]\big), y_i\right)
\label{eq:F}
\end{align}

\begin{proposition}~\label{prop:0}
Let $F$ and $\loss$ are defined in Eqs.~\ref{eq:F} and~\eqref{eq:loss}, respectively. Then, we have:
\begin{align}
&\textstyle\min_{\set{\ucal{i}, \vcal{i}}: |\ucal{i}\cp\vcal{i}| \le q_{\max}} \sum_{i\in D_i}\loss(h,p; \ucal{i}, \vcal{i}\given \ocal{i} ) \nn\\
& \qquad \qquad \le \textstyle \min_{\ucal{b}: |\ucal{b}| \le q_{\max}}\sum_{i\in D_b}F(h,p; \ucal{b} \given \ocal{i} ) 
\end{align}
\vspace{-3mm}
\end{proposition}

The set-optimal value of the above objective is an upper bound of  $\loss ()$  in Eq.~\eqref{eq:loss}, as stated formally here~\footnote{\scriptsize Proofs of all technical results are  in Appendix~\ref{app:proofs}}.
Hence, we instead of minimizing $\loss$, 
 we seek to solve the following optimization problem for each bucket $b$.
\begin{align}
\min_{\theta_b,\phi_b, \ucal{b} } \sum_{i\in D_b}F(h_{\theta_b},p_{\phi_b};\ucal{b} \given \ocal{i}), \ \text{s.t., } |\ucal{b}| \le q_{\max} \label{eq:objxx}
\end{align}
The objective $\loss(\cdot)$~\eqref{eq:loss} queries two different sets of features from the oracle and the generator, \ie, $\ucal{b}\cp\vcal{b}$ and $\vcal{b}$. In contrast, $F$ queries the same set of features $\ucal{b}$ from both oracle and the generator.  Here, it assigns more weights to the loss from the generated features (oracle features) if the pre-trained classifier is less (more) uncertain from the generated features. In the absence of the generator, $F$ only contains the loss for the oracle features, \ie, $F(h_{\theta_b},p_{\phi_b};\ucal{b} \given \ocal{i})= \ell(h_{\theta_b} (\xb_i [\ocal{i} \cup \ucal{b}]),y)$ and the task reduces to the well-known feature selection problem in~\cite{elenberg2018restricted}.
% Next, we formally state the relationship between  $F(\cdot)$ and   $\loss(\cdot)$\footnote{\scriptsize Proofs of all technical results are  in Appendix~\ref{app:proofs}}.

\xhdr{Optimization over $\vcal{b}$} The above optimization involves only $\ucal{b}$ as the optimization variables, and is independent of $\vcal{b}$. Once we compute $\theta_b ^*, \phi_b ^*, \ucal{b} ^*$, \ie, the solution of the optimization~\eqref{eq:objxx}, we use them to compute the set $\vcal{b}$ by solving the following optimization problem:
\begin{align}
&\hspace{-3mm} \min_{\vcal{b}: |\vcal{b}|\le\lambda } \sum_{i\in D_b} (1-\Delta_i(\vcal{b}) )\cdot \loss(h_{\theta_b ^*},p_{\phi_b ^*}; \ucal{b}^*, \vcal{b}\given \ocal{i} ) \label{eq:objyy}
 \end{align}
 % \EE_{\xb'\sim p_{\phi^* _b}(\bullet \given \xb[\ocal{i}])}
where $\lambda$ is a hyperparameter.

\xhdr{Objectives in ~\eqref{eq:objxx},~\eqref{eq:objyy} as set functions} Here, we represent the objectives in the optimizations~\eqref{eq:objxx},~\eqref{eq:objyy} as set functions, which would be later used in our training and inference methods and approximation guarantees. Given $U$,  the optimal solution of the objective $\theta_b ^*(\ucal{}), \phi_b ^*(\ucal{})$  minimizes 
  $\sum_{i\in D_b}F(h_{\theta_b},p_{\phi_b};\ucal{} \given \ocal{i})$. Thus,  $\min_{\theta_b, \phi_b}\sum_{i\in D_b}F(h_{\theta_b},p_{\phi_b};\ucal{} \given \ocal{i})$ becomes a set function in terms of $\ucal{}$. On the other hand,  $  \sum_{i\in D_b} (1-\Delta_i(\vcal{}) )\cdot \loss(h,p; \ucal{b}^*, \vcal{}\given \ocal{i} )$ is also a set function in terms of  $\vcal{}$.
 To this end, we define $\gf$ and $\gl$ as follows 
 given any set $\ucal{}$ we  
 describe the optimal value of this training problem as the following set function:
 \begin{align}
&   \hspace{-2mm}  \gf(\ucal{})  = \hspace{-1mm}\sum_{i\in D_b}F(h_{\theta_b ^*(\ucal{})},p_{\phi_b ^*(\ucal{})};\ucal{} \given \ocal{i}),  \label{eq:Gdefx}
 \\
 &\hspace{-2mm} \gl(\vcal{})  = \hspace{-1mm}\sum_{i\in D_b} (1-\Delta_i(\vcal{}) )  \loss(h_{\theta_b ^* },p_{\phi_b ^* }; \ucal{b}^*, \vcal{}\given \ocal{i} )
     \label{eq:Gdef}
 \end{align}

 \subsection{Training and inference algorithms}

 % \xhdr{Characterization of objectives~\eqref{eq:objxx} and~\eqref{eq:objyy}}

% In practice, we observed that greedy algorithm works well to solve the optimization~\eqref{eq:objxx} and~\eqref{eq:objyy}. This leads us to investigate the set function centric properties of these problems.
 \begin{figure}[t]
\vspace{-9mm}
 \begin{minipage}{0.46\textwidth}
\hrule
%%%
\small
\begin{algorithmic}[1]
    \FUNCTION{\textsc{GreedyForU}($q_{\max},b, h,p$)}
   % \STATE  \REQUIRE Access to computation of $G_F$ and $F$
    \STATE $U_b \leftarrow \emptyset$, 
    \STATE $\theta_b  \gets \textsc{Train}(h, D_b,\set{\ocal{i}})$  \texttt{\small \#pretraining}
      \STATE $\phi_b \hspace{-0.5mm} \gets \textsc{Train}(p, D_b,\set{\ocal{i}})$ \texttt{\small\#pretraining}
        \FOR{ iter $\in [q_{\max}]$ }
          \STATE   \texttt{\small \# Use $h_{\theta _b }$, $p_{\phi _b}  $ to compute $\gf,F$ }
             \STATE $e^* \gets {\argmin}_{e\not\in \ucal{b}\cup \ocal{i}}\ \gf(e\given \ucal{b}) $
             \IF{ $\gf(e^*\given \ucal{b})  < 0$} 
             \STATE $U_b \gets U_b \cup e^*$ 
              \STATE $\theta _b , \phi _b  \gets \textsc{Train}(F,\ucal{b},b)$
              \ELSE
              \STATE \textbf{break}
              \ENDIF
         \ENDFOR
        % \STATE $e^* \leftarrow \underset{e}{\mathrm{argmin}}\sum_{i\in b}F(h_{\theta_b},p_{\phi_b};\ucal{b}\cup\{e\} \given \ocal{i}); ~e \in E $
        \STATE \textbf{Return} $U_b, \theta _b , \phi _b $
    \ENDFUNCTION
        \end{algorithmic}
\begin{algorithm}[H]
\small
\caption{Inference}
\begin{algorithmic}[1]
    \REQUIRE Observed test feature, $\xb[\ocal{}]$, threshold $\tau$, trained models $h_{\theta^* _b}, p_{\phi^*_b}$, $\set{\ucal{b} ^*,\vcal{b}^*}$.
    \STATE $b \leftarrow$ \textsc{FindBucket}$(\xb^*[\ocal{}])$
    \STATE Query $\xb[\ucal{b}^*\cp \vcal{b}^*]$ from oracle
    \STATE $\xg[\vcal{b}^*]\sim p_{\phi^* _b}(\bullet \given \xb^*[\obs\cup  \ucal{b}^* \cp \vcal{b}^*])$ 
    \STATE $\xb_{\text{all}}  = \xb[\obs\cup  \ucal{b} ^*\cp \vcal{b}^*]\cup \xg[\vcal{b}^*]$
    % \STATE Query $\ucal{b}\cp \vcal{b}$
    \IF{$\max_{y} h_{\theta^* _b }(\xb_{\text{all}})[y]< \tau$}
        \STATE Query $\xb[\vcal{b}^*]$ from oracle,
        $\xb_{\text{all}}  = \xb[\obs\cup  \ucal{b}^*]$
    \ENDIF
         \STATE $y^* \leftarrow \argmax_{y} h_{\theta^* _b }(\xb_{\text{all}})[y]$ 

    \STATE \textbf{Return} $y^*$
\end{algorithmic}
\label{alg:inference}
\end{algorithm}
\end{minipage}
\end{figure}
% Hence, we resort to extensions of 
% existing properties~\cite{mualem2022using,elenberg2018restricted} and assumptions on the objectives.
% We introduce new extensions of partial monotonicity~\cite{mualem2022using} and weak submodularity,
% and provide novel characterizations of the objectives in the lens of these extensions. 
% However, characterization of the underlying set functions is challenging in generic setting. To t 
% \footnote{\scriptsize $\theta_b ^*(\ucal{}), \phi_b ^*(\ucal{})$ denotes as the optimal solution of~\eqref{eq:objxx} for fixed $\ucal{}$, whereas $\theta_b ^*, \phi_b ^*$ denote the solution of the joint optimization~\eqref{eq:objxx}, when $\ucal{b}$ is also optimized, \ie,  $\theta_b ^* = \theta_b ^*(\ucal{b} ^*)$, $\phi_b ^* = \phi_b ^*(\ucal{b} ^*)$}

% We first provide two set
\xhdr{Training}
Our training algorithm uses greedy heuristics combined with model training to solve the optimization problems~\eqref{eq:objxx} and~\eqref{eq:objyy} for computing $\theta_b ^*, \phi_b ^*, \ucal{b} ^*, \vcal{b}^*$. Given the dataset $D$ and the number of buckets $B$, we first partition the dataset into $B$ buckets (\textsc{Partition}($\cdot$)) and perform pre-training of $h_{\theta_b}$ and $p_{\phi_b}$ 
on $D_b$ based on the observed features $\set{\ocal{i}}$. For the generator, we use a $\beta-$VAE ~\cite{higgins2017beta} to optimize regularized ELBO.
Then, for each bucket $b$, we leverage two greedy algorithms for non-monotone functions~\cite{mualem2022using,harshaw2019submodular},  one for computing $\ucal{b}$ (\textsc{GreedyForU}) and the other for $\vcal{b}$ (\textsc{GreedyForV}). At each iteration, \textsc{GreedyForU} keeps adding a new feature $e=e^*$ to $\ucal{b}$ which minimizes $\gf(e\given\ucal{b})$ as long as it admits a negative marginal gain, \ie, $\gf( e^* \given \ucal{b})< 0$ and $|\ucal{b}| \le q_{\max}$. Similarly, we use the greedy algorithm on $\gl$ to get $\vcal{b}$. \textsc{GreedyForU} needs to train the model for every candidate for each new element. To tackle this, in practice we adopt three strategies. (1) We directly use the model parameters $\theta_b , \phi_b $ obtained in the previous step  to compute $\gf$ during search of the potential new candidates $e$ (line 5), without any new training. After we select $e^*$, we perform two iterations of training. Since we consider adding only one element, such a small amount of training is enough, beyond which we did not see any observable improvement.
(2) We tensorize the operation $\argmin_e \ G_{\bullet}(e\given\ucal{b})$ instead of enumerating $G_{\bullet}$ for all candidates $e$. Note that we utilize all the features available in the training set during training. However, this does not amount to cheating, as in the inference algorithm, only the required features are queried. Such a protocol is widely followed in works including \cite{acflow} and \cite{eddi}, helping them to generalize to the set of all features.
% In practice, the last $\lambda$ features of $\ucal{b}$ may be used as $\vcal{b}$ as the classifier is least sensitive to their noise.
%  \begin{wrapfigure}[11]{r}[-115pt]{0.2\columnwidth}
% \vspace{-9mm}
   
% \end{wrapfigure}

\xhdr{Inference} Given a test instance with a subset of observed features $\xb[\ocal{}]$, we first find the bucket $b$.
It then queries the oracle for $\xb[\ucal{b}\cp \vcal{b}]$. Taking $\xb[\obs \cup \ucal{b}\cp \vcal{b}]$ as input, the generator produces $\xg[\vcal{b}]$. 
% The algorithm then evaluate $h_{\theta_b}\big(\xb^* [\obs\cup  \ucal{b} \cp \vcal{b} ] \cup \xg [\vcal{b}]\big)$. 
If the confidence of the classifier with these features  is lower than a threshold $\tau$, then \our\ does not use the generated features for the final prediction and, further query  $\xb[\vcal{b}]$
from the oracle and predict $y$ using $\xb[\ocal{}\cup \ucal{b}]$.
However, otherwise if the confidence of the classifier with the features is higher than $\tau$, we use the generated features and compute $y$ using $\xb[\obs\cup  \ucal{b} \cp \vcal{b}]\cup \xg[\vcal{b}]$.

 \subsection{Characterization of our optimization tasks}
 Here, we present a set function based characterization of our objectives~\eqref{eq:objxx} and~\eqref{eq:objyy} (or, equivalently,~\eqref{eq:Gdefx} and~\eqref{eq:Gdef}), beginning with a discussion on hardness analysis. Then, we use those characterizations to prove the approximation guarantee of our algorithms. 
 
\xhdr{Hardness} At the outset, our goal is to first compute the optimal $\ucal{} = \ucal{}^*$ by minimizing $\gf(\ucal{})$
and then use this $\ucal{}^*$ to compute the optimal $\vcal{}$ by minimizing $\gl(\vcal{})$.
The optimization of $\gf(\ucal{})$ ----- or, equivalently the optimization~\eqref{eq:objxx}--- is a discrete continuous optimization problem, since it involves model training in conjunction with subset selection. 
 Given $U$, one can find the optimal solution of the objective $\theta_b ^*(\ucal{}), \phi_b ^*(\ucal{})$ in polynomial time when the objective is convex with respect to both $\theta$ and $\phi$. 
However, the simultaneous computation of
the optimal set $\ucal{} ^*$ and the model parameters $\theta_b ^*$ and $\phi_b ^*$
is NP-Hard even in simple cases, \eg,  
sparse feature selection~\cite{elenberg2018restricted}.

\xhdr{Set function centric characterizations}
We first extend the   notions of  partial monotonicity~\cite{mualem2022using}
and $\gamma$-weak submodularity~\cite{elenberg2018restricted,harshaw2019submodular}.
\begin{definition}
 Given a set function $G:2^{[n]} \to \RR $, two sets $S$ and $T$ with $S,T \subseteq [n]$ and the marginal gain $G(S\given T):= G(S\cup T) - G(T)$. Then we define the following properties.\\
 % \textbf{ (1)} $m$-Partial monotonicity:
 % The set function $G$ is $m-$ partially monotone  if $\frac{G(T)}{G(S)} \in [m,\infty]$ with $m\ge 0$ for all $S,T$ with $S\subseteq T \subseteq [n]$ \cite{mualem2022using}. 
\textbf{ (1)} $(\mmin,\mmax)$-Partial monotonicity: The set function $G$ is $(\mmin,\mmax)$-partially monotone ($\mmin \ge 0$)  if $\frac{G(T)}{G(S)} \in [\mmin,\mmax]$ for all $S,T$ with $S\subseteq T \subseteq [n]$. 
 \textbf{ (2)} $(\gmin,\gmax)$-Weak submodularity: The set function $G$ is $(\gmin,\gmax)$-weakly submodular  if $\frac{\sum_{u\in S}G(u\given T)}{|G(S\given T)|} \in [\gmin,\gmax]$ for all $S,T$ with $S\cap T = \emptyset$.
% \vspace{-2mm}
\end{definition}
Similar to~\citet{mualem2022using}, we define that $G(T)/G(S)=1$ if $G(S) = 0$. Note that,  a $(\mmin,\mmax)$ partially monotone function $G$ is monotone increasing (decreasing) if $\mmin = 1\, (\mmax =1)$. Moreover,  $m-$partial monotonicity introduced in~\cite{mualem2022using} implies $(m,\infty)$-partial monotonicity. A $\gamma-$weakly submodular function~\cite{elenberg2018restricted,harshaw2019submodular} is $(\gamma,\infty)$-weakly submodular. Next, we assume  boundedness of few quantities, allowing us to characterize $\gf$ and $\gl$.
\begin{assumption} 
\label{assumption}
\textbf{(1)} Bounded difference between uncertainties across two feature subsets:
Given a bucket $b\in [B]$, $|\Delta_i(\ucal{}) - \Delta_i(\vcal{})| \le \ee_{\Delta}$.
\textbf{(2)} Bounds on uncertainty and loss: 
$0< \Delta_{\min} \le |\Delta_i(\ucal{})| \le \Delta_{\max} $.
$ 0< \ell_{\min} \le |\ell(h_\theta(\xb_i),y_i)| \le \ell_{\max}$.
\textbf{(3)} Lipschitzness: The loss function $\ell(h_\theta(\xb),y)$ is Lipschitz with respect to $\xb$. \textbf{(4)} Boundedness of features: $||\xb_i||$, $||\xg_i|| \le \ee_x$, for all $i$. 
In Appendix~\ref{app:assump}, we discuss the validity of these assumptions as well as present the values of $\ee_\bullet$ across different datasets.
\end{assumption}
% Next, we leverage the above assumption to state our first result on partial monotonicity.
\begin{theorem}[$(\mmin,\mmax)$-Partial monotonicity]
\label{thm:mon}
\textbf(1) The set function $\gf$  is $(\mmin_F,\mmax_F)$-partially monotone where  $\mmax_F = 1+ {K}_F\ee_x \ee_{\Delta }$ and $\mmin_F = \left(1+ {K}_1\ee_x +  {K}_2 \ee_{\Delta } +  {K}_3\right)^{-1}$. 
\textbf(2) The set function $\gl$  is
$(\mmin_\loss,\mmax_\loss)$-partially monotone where  $\mmax_\loss = 1+ {K}_\loss\ee_x$ and $\mmin_\loss = 
( 1+ {K}_\loss\ee_x)^{-1}$. 
Here ${K}_{\bullet}$ depend on the Lipschitz constant of the loss with respect to $\xb$ and the bounds  on loss $\ell$ and the uncertainty $\Delta$.  
\vspace{-2mm}
\end{theorem}
Partial monotonicity of $\gf$ suggests that if the variation of uncertainty across different feature sets goes small ($\ee_{\Delta} \to 0$) or the generator is extremely accurate  ($\ee_x \to 0$), then we have: $\mmax \to 1$, meaning that $\gf$ is monotone decreasing.
If we put $\Delta_i(\ucal{b}) = 1$ for all $i\in D_b$ in the expression of $F$in Eq.~\eqref{eq:F}, then the optimization~\eqref{eq:objxx} becomes  oblivious to the generative model $p_{\phi_b}$. 
In such a case,  $\gf$ is monotone decreasing since, $\ee_\Delta = 0$. This result coincides with the existing characterizations of the traditional feature selection problem~\cite{elenberg2018restricted}. 
In the context of the optimization for $V$, note that if the generator is very accurate $(\ee_x \to 0)$, then Partial monotonicity of $\gl$ implies that $\gl(V) $ is almost constant for all $V$. In such a case, one can use the feature generator to generate the entire set  $\vcal{b}=\ucal{b}^*$ and save the entire budget $q_{\max}$. Since, we have $\gl(S \given T) = 0$ in a simple cases where $\ee_x $ or $\ee_\Delta = 0$, $(\gmin,\gmax)$-weak submodularity does not hold for $\gl$ in most cases. Thus, we present the $(\gmin,\gmax)$-weak submodularity property of only $\gf$ under additional assumptions.
\begin{theorem}[$(\gmin,\gmax)$-weak submodularity] Assume that the loss $\ell(h_{\theta}(\xb),y)$  be convex in $\theta$ with
$\nabla_{\theta}\ell(h_{\theta}(\xb),y) \le \nabla_{\max}$ 
and $\text{Eigenvalues}\set{\nabla^2 _{\theta} \ell(h_{\theta}(\xb[S]),y)} \in [\zeta_{\min}, \zeta_{\max}]$ for all $S$; 
% the parameters of the generator 
% $\phi$ is bounded, \ie, $||\phi|| \le \phi_{\max}$; $\Delta_{\max} > \max\set{0,1-{\nabla^2 _{\max}}/{4\zeta_{\max} \phi_{\max} 
% L_{\phi}}} $ and $\ee_{\Delta} \ee_x < {\nabla^2 _{\max}}/{2\zeta_{\max}L_{x}} - {(1-\Delta_{\max}) \phi_{\max} L_{\phi} }/{L_{x}} $ 
Then,
\label{thm:sub}
the set function $\gf$  is $(\gmin_F,\gmax_F)$-weakly submodular 
with
\begin{align}
 \hspace{-2mm}\overline{\gamma}_F \le \max&\left\{  \frac{   -\frac{\nabla_{\max} ^2}{2\zeta_{\max}}+ K_{\gmax,1} L_\phi  + K_{\gmax,2} \ee_{\Delta} \ee_x  }{\frac{\nabla_{\max} ^2}{2\zeta_{\min}}+ K_{\gmax,3} L_\phi   + K_{\gmax,4}  \ee_{\Delta} \ee_x   }, \right. \nn\\[-1ex]
& \qquad \ \left.  \frac{    -\frac{\nabla_{\max} ^2}{2\zeta_{\max}}+ K_{\gmax,1} L_\phi  + K_{\gmax,2} \ee_{\Delta} \ee_x  }{\frac{\nabla_{\max} ^2}{2\zeta_{\max}}- K_{\gmax,5}L_\phi -K_{\gmax,6} \ee_{\Delta} \ee_x  }
 \right\}; \nn\\
 \gmin_F \ge   
& \frac{   -\frac{\nabla_{\max} ^2}{2\zeta_{\min}}- K_{\gmin,1} L_\phi - K_{\gmin,2} \ee_{\Delta} \ee_x  }{\frac{ \nabla_{\max} ^2}{2\zeta_{\max}}- K_{\gmin,3} L_\phi  -  K_{\gmin,4} \ee_{\Delta} \ee_x    }
\end{align}
where $K_{\gmax,\bullet}$ and $K_{\gmin, \bullet}$ are constants that depend on the Lipschitz constant of the loss w.r.t. $\xb$; $L_{\phi}$ is the Lipschitz constant of $F$ w.r.t. $\phi$.
\end{theorem}
In the absence of the generator, $\ee_\Delta, \ee_x$, $L_{\phi}$ are zero. Then, $-\gf$ is $\gamma$-weakly submodular with $\gamma > \zeta_{\min}/\zeta_{\max}$ which coincides with the results of~\cite{elenberg2018restricted}.
Next, we present the approximation guarantee of our greedy algorithm for solving the optimization problem~\eqref{eq:objxx} presented via $\textsc{GreedyForU}$, when $\ell(\hb_{\theta} (\xb),y)$ is convex in $\theta$. 
\begin{theorem} Assume that, given a bucket $b$,  $\gf$ is  $(\mmin_F,\mmax_F)$-partially monotone, $(\gmin_F,\gmax_F)$-weakly submodular set function. Suppose $\ucal{b}^*$ is
the output of $\textsc{GreedyForU}$ in Algorithm~\ref{alg:training}. Then, if $OPT=\argmin_{U_b: |U_b|\le{q_{\max}}} \gf(U_b)  $, we have 
\begin{align}
\hspace{-2mm}\gf(\ucal{b}^*) & \le   m_F \gf(OPT)\nn\\ 
& \  -\left(1-\frac{\gamma_F}{q_{\max}} \right)^{q_{\max}} \Big(m_F \gf(OPT)-\gf(\emptyset)\Big) \nn
\end{align}
where $m_F=\max\left(\mmax_F, 2\mmax_F/\mmin_F \right)$ and $\gamma_F = \max(\gmax_F,-\gmin_F)$.
\label{thm:greedy}
\vspace{-1mm}
\end{theorem}
We discuss the quality of the approximation guarantees for different datasets in our experiments in Appendix~\ref{app:assump}.

\newcommand{\dpr}{DP}
\newcommand{\mnist}{MNIST}
\newcommand{\cifar}{CIFAR100}
\newcommand{\ti}{TinyImagenet}

% \vspace{-}
\section{Experiments}
\label{sec:exp}
% In this section, we experiment with four real datasets and show that our method outperforms several baselines.
% \subsection{Experimental setup}

\xhdr{Datasets}
We experiment with four datasets for the classification task; \dpr\ (disease prediction), \mnist, \cifar\ and \ti\ (TI). 
For \dpr, the number of features $n=132$ and the number of classes $|\Ycal|=42$. Here, different features indicate different symptoms and the classes indicate different diseases. For the image datasets, we take the pixels or groups of pixels as the features, following previous work~\cite{acflow,gsm}.
Further details about the datasets are provided in Appendix~\ref{app:setup}. 
For each dataset, the average number of observed features $\EE[|\ocal{i}|] \approx n/10$. 
% The number of $\ocal{i}$ per each 
% instance  $\overline{N_{\ocal{}}}$ varies across datasets. 
% For \dpr, \mnist, \cifar\ and \ti, $\overline{N}_{\ocal{}} =20, 10,5$ and $5$ respectively. Across all datasets, the size of $|\ocal{i}| \approx n/10$, where $n$ is the total number of features.

\xhdr{Baselines}
We compare \our\ with six state-of-the art baselines: JAFA \cite{jafa}, EDDI \cite{eddi}, ACFlow \cite{acflow}, GSM \cite{gsm}, CwCF \cite{cwcf} and DiFA \cite{difa} each
with two variants  (batch and sequential).  
In the first variant, we deploy these methods to perform in the batch setting. Specifically, we use top-$q_{\max}$ features provided by the feature selector in batch, instead of querying the oracle features one by one.  This feature selector is a policy network in RL methods~\cite{gsm,jafa,cwcf,difa}, greedy algorithms for maximizing rewards in~\cite{eddi,acflow}. 
Note that our approach uses mixture models on the partitioned space, which enhances the expressivity of the overall 
system. To give the baselines a fair platform for comparison, we deploy a mixture of classifiers on the same partitioned space, where the models are trained using the observed features and the corresponding queried features of the baselines. \todo[disable]{Need to mention that in some places we do use the off the shelf thing. Also too aggressive text against the raw baselines.} This is done in the cases where, off-the-shelf use of their methods led to poor performance, especially in the larger datasets (Appendix~\ref{app:exp}). 
In the sequential variant, we make the baselines to acquire features sequentially.
Since our goal is to acquire features in batch, the first setting gives a fairer platform for comparison across all methods. Thus, we only present the results of the batch setting in the main and, defer the results of the sequential setting in Appendix~\ref{app:exp} 
% Here, they acquire one feature at a time and acquire the next feature based on the values of features acquired thus far. We defer the results of this variant to Appendix~\ref{app:exp}. 

% In the second variant, we modify these methods to perform in the batch setting. Specifically, we use top-$q_{\max}$ features provided by feature selector in batch, instead of querying the oracle feature one by one.  This feature selector is a policy network in RL methods~\cite{gsm,jafa,cwcf,difa}, greedy algorithms for maximizing rewards in~\cite{eddi,acflow}.  Note that our approach uses mixture models on the partitioned space, which enhances the expressivity of the overall 
% system. To give the baselines a fair platform for comparison, we deploy a mixture of classifiers on the same partitioned space, where the models are trained using the observed features and the corresponding queried features of the baselines. \todo[disable]{Need to mention that in some places we do use the of-the-shelf thing. Also too aggressive text against the raw baselines.} This is done in the cases where, off the shelf use of their methods led to poor performance, especially in the larger datasets (Appendix~\ref{app:exp}).

\xhdr{Classifier $h$ and the generator $p$} For \dpr\ and \mnist, the classifier $h$ consists of  linear, ReLU  and linear networks, cascaded with each other, with hidden dimension $32$.
For \cifar, $h$ is WideResnet~\cite{zagoruyko2016wide} for \ti, $h$ is EfficientNet ~\cite{tan2021efficientnetv2}. We use the same classifier $h$ across all baselines too.
The generator $p$ is a variational autoencoder. It contains an encoder and a decoder that are pre-trained on the observed instances as a $\beta$-VAE~\cite{higgins2017beta}. Details are provided in Appendix~\ref{app:setup}. We set the number of buckets $B=8,8,4,4$  for DP, MNIST, CIFAR100 and \ti\ respectively, which we set using cross validation. 
%%
% For \dpr\ and \mnist, we use a transformer model as the encoder. Given $\xb_i[\obs_i]$, the encoder produces a set embedding $\zb_{i}$ of shape $|\obs_i| \times d_{\zb}$ using pretrained position embeddings, where we set $d_{\zb} = 20$ for \dpr\ and $d_{\zb} = 69$ for \mnist. The decoder then takes the position wise concatenation of $\xb$ and $\zb$, $\text{Concat}\left([\xb_i[\text{pos}],\zb[\text{pos}]]\right)_{\text{pos} \in \obs_i}$ as input. The decoder is a cascaded network of linear, ReLU and linear layer. For \cifar\ and \ti\ datasets, we use Resnet-152 encoder~\cite{he2016deep}. The input image is masked with the set of observed features and fed to the encoder. Here, the decoder uses transposed convolution and is borrowed from the generator of DCGAN~\cite{radford2015unsupervised}. 

\xhdr{Evaluation protocol} We split the entire dataset in 70\% training, 10\% validation and 20\% test set. We use the validation set to cross validate $\lambda$ and the number of buckets $B$. Having observed a feature $\xb[\obs]$ during test, we use the learned method to compute the set of new features to be acquired $\unobs$ and $\vcal{}$ for a given budget $q_{\max}$. We use $\xg[\vcal{}]$
by drawing from the generator, whereas we query $\xb[\unobs\cp \sub]$ from the oracle. Finally, we feed all the gathered feature into the classifier and compute the predicted label $\hat{y}$. We cross validate our results $20$ times to obtain the p-values.

\subsection{Results}
% \vspace{-1mm}
\begin{figure*}[t]
    \centering
    \includegraphics[width=0.45\linewidth]{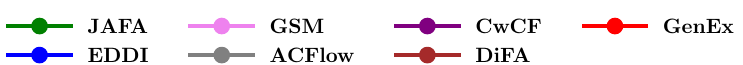}\\
\subfloat[DP]{\includegraphics[width=.18\textwidth]{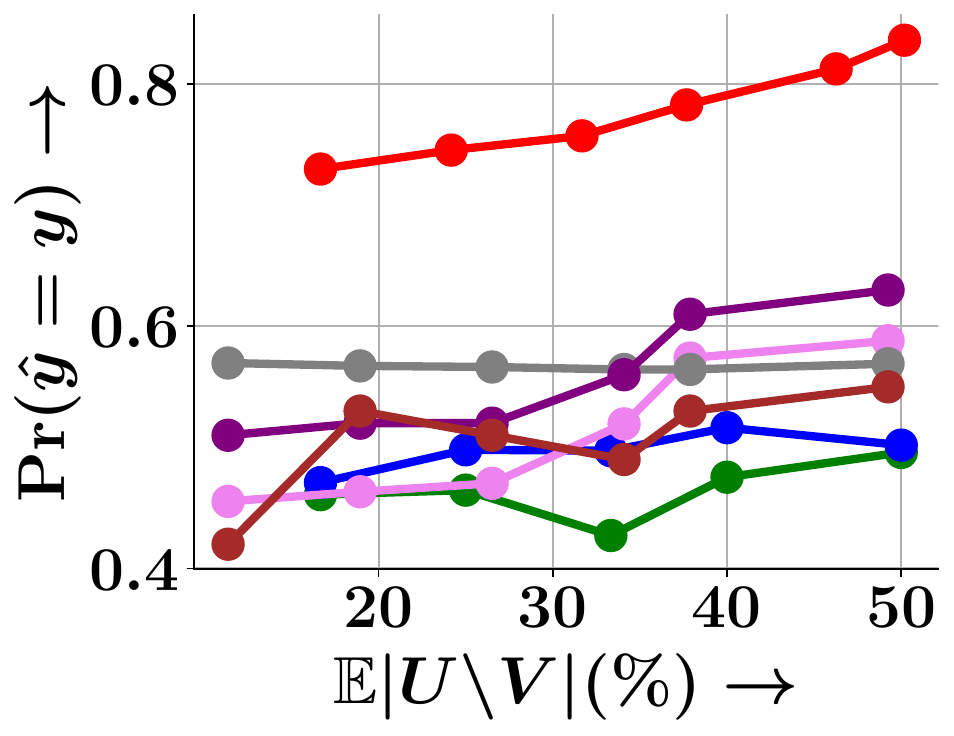}} \hspace{1mm}
\subfloat[MNIST]{\includegraphics[width=.17\textwidth]{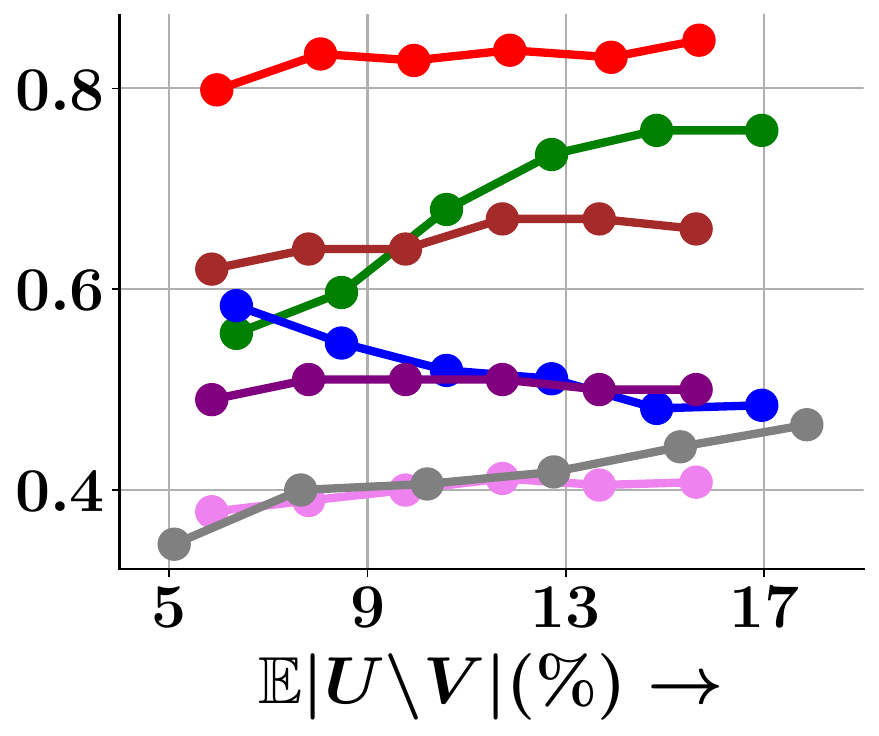}}\hspace{1mm}
\subfloat[CIFAR100]{\includegraphics[width=.17\textwidth]{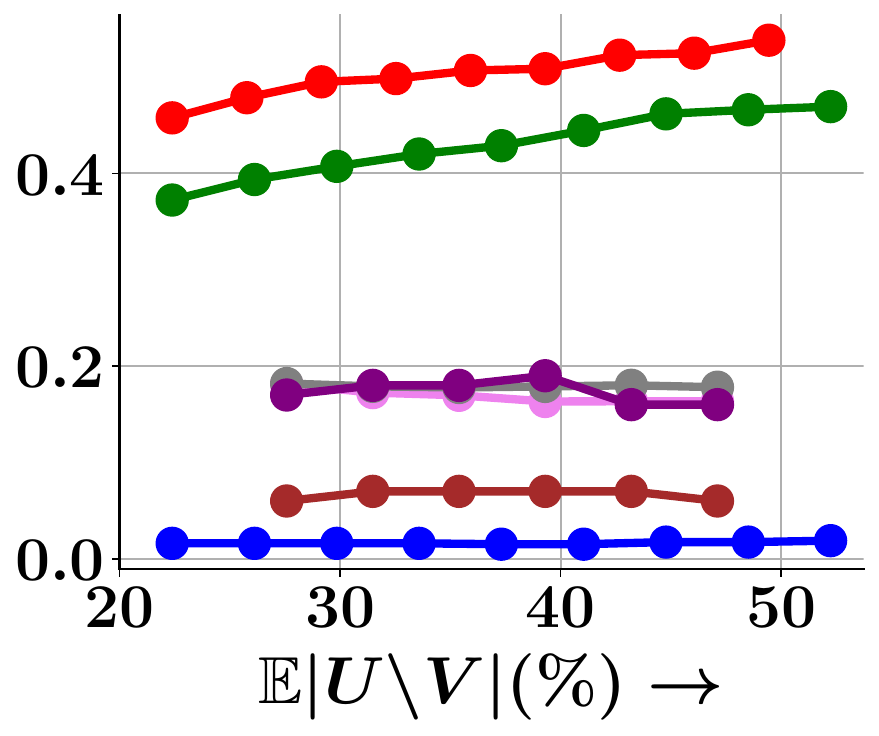}}\hspace{1mm}
\subfloat[TinyImageNet]{\includegraphics[width=.17\textwidth]{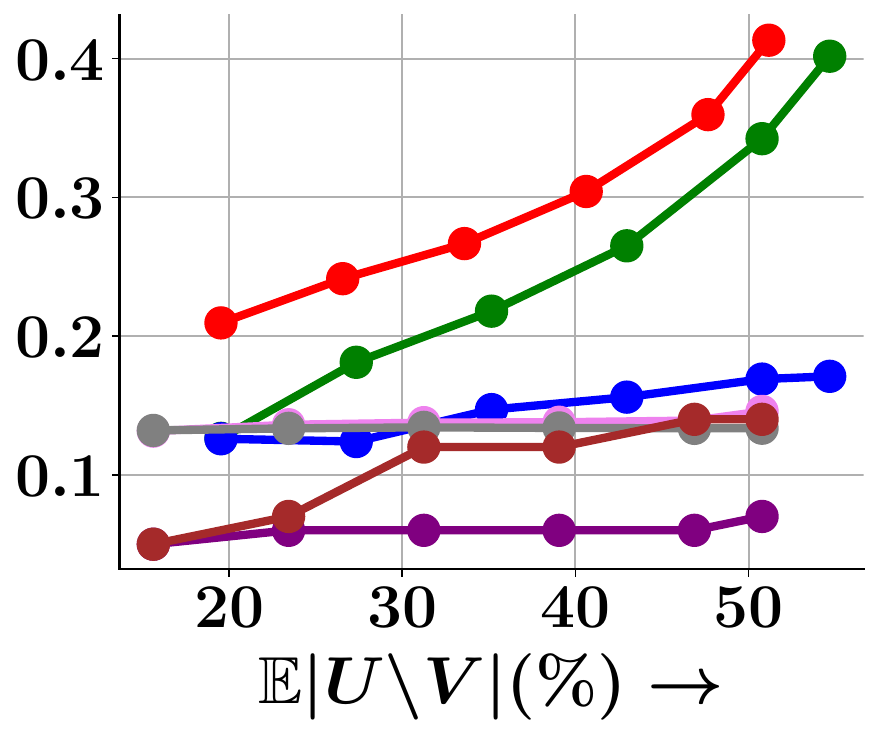}}
% \vspace{-1mm}
\caption{Comparison  of \our\ against batch variants of baselines, \ie, JAFA \cite{jafa}, EDDI \cite{eddi}, ACFlow \cite{acflow}, GSM \cite{gsm}, CwCF \cite{cwcf} and DiFA \cite{difa} in terms of the classification accuracy varying over the average number of oracle queries $\EE|\ucal{} \cp \vcal{}|$, for all four datasets.}
% \vspace{-1mm}
\label{fig:main}
\end{figure*}
 \xhdr{Comparison with baselines}
 First, we compare the prediction accuracy of \our\ against 
 % both the sequential and batch variants of
 the baseline models for different 
 value of the maximum permissible number of oracle queries $q_{\max}$ per instance. The horizontal axis indicates $\EE[|V|/|U|]$, the average number of oracle queries per instance.
Figure \ref{fig:main} summarizes the results. 
We observe:
\textbf{(1)} \our\ outperforms all these baselines by a significant margin. 
% including RL based methods. We directly attempt to solve the combinatorial problem, whereas RL policies often approximate the search using neural networks.
% We have a combinatorial greedy algorithm which, unlike RL policies, uses a surrogate for the actual objective taking generated features into account.
The competitive advantage provided by \our\ is statistically significant (Welch's t-test, p-value $< 10^{-2}$). \textbf{(2)} JAFA performs closest to ours in large datasets. \textbf{(3)} The baselines are not designed to scale to a large number of features, as asserted in the classification experiments in \cite{gsm}, and hence their accuracy stagnates after acquiring a few features~\cite[Fig. 6, 7]{gsm} and~\cite[Fig. 3]{jafa}. 
% We compare the baselines on the full feature space, while many of the baselines are compared on reduced feature spaces in their own papers \cite{gsm, difa}.
% \textbf{(3)} JAFA performs  closest to \our. It has an encoder network which is shared across both classifier and the policy network. This allows for information sharing, as mentioned in their paper~\cite[Section 4]{jafa}

 \begin{figure}[!t]
    \centering
 \hspace{-2mm}   \begin{minipage}{0.40\textwidth}
        \centering
% \includegraphics[width=0.9\linewidth]{FIG/legend_appendix.pdf}
%  \\[-3ex]
 \subfloat{\includegraphics[width=.47\textwidth]{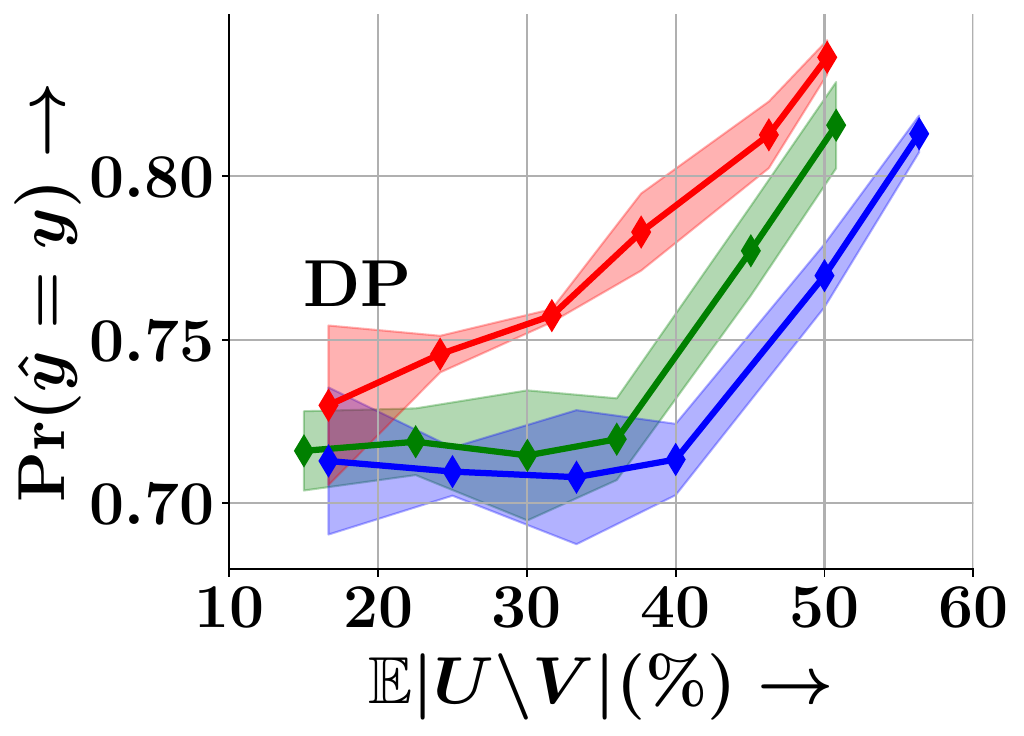}} \hspace{4mm}
\subfloat{\includegraphics[width=.43\textwidth]{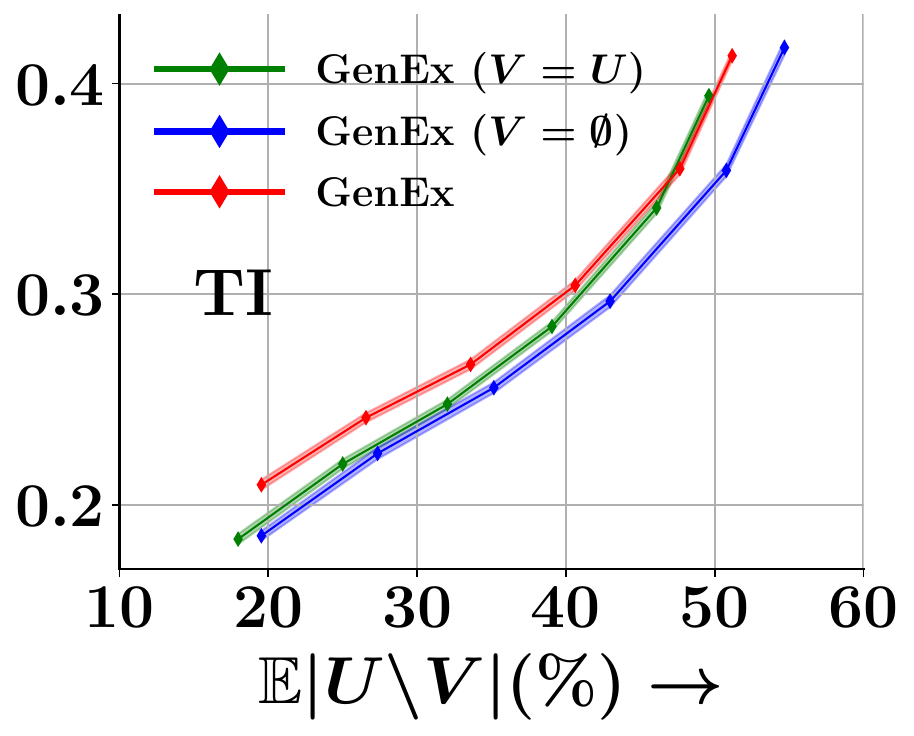}} \\[-1ex]
      \caption{\small Ablation study: $\Pr(\hat{y} = y)$ vs. $\EE[|U\cp V|]$}
        \label{fig:ablation1}
    \end{minipage} \\[2ex]
        \begin{minipage}{.40\textwidth}
        \centering
 \subfloat{\includegraphics[width=.46\textwidth]{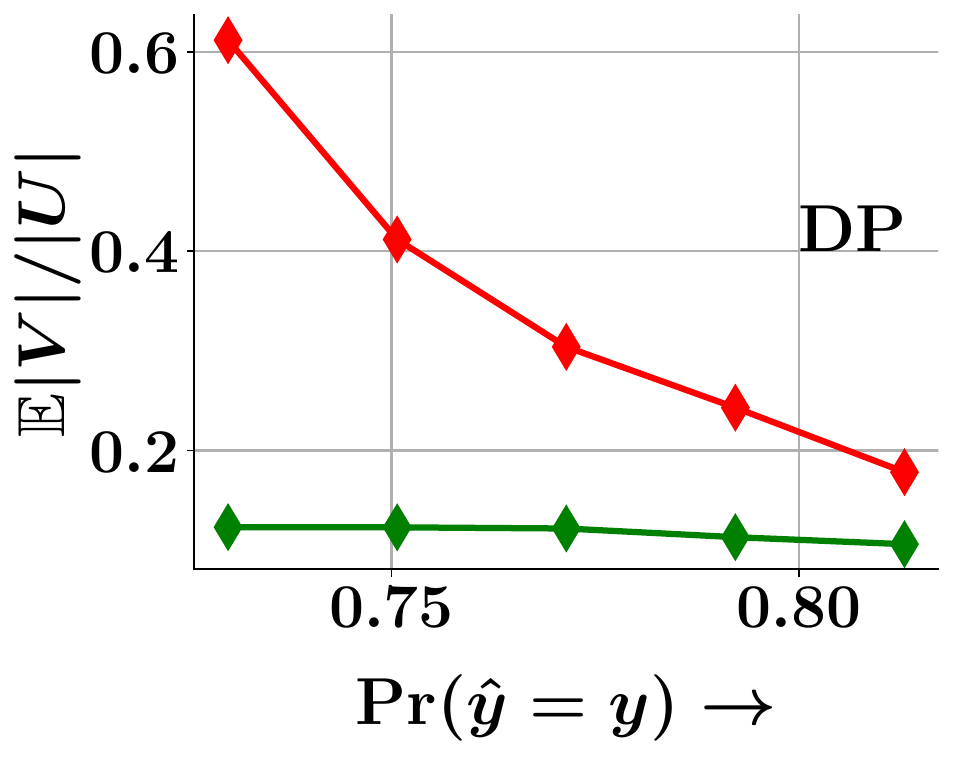}} \hspace{3mm}
\subfloat{\includegraphics[width=.46\textwidth]{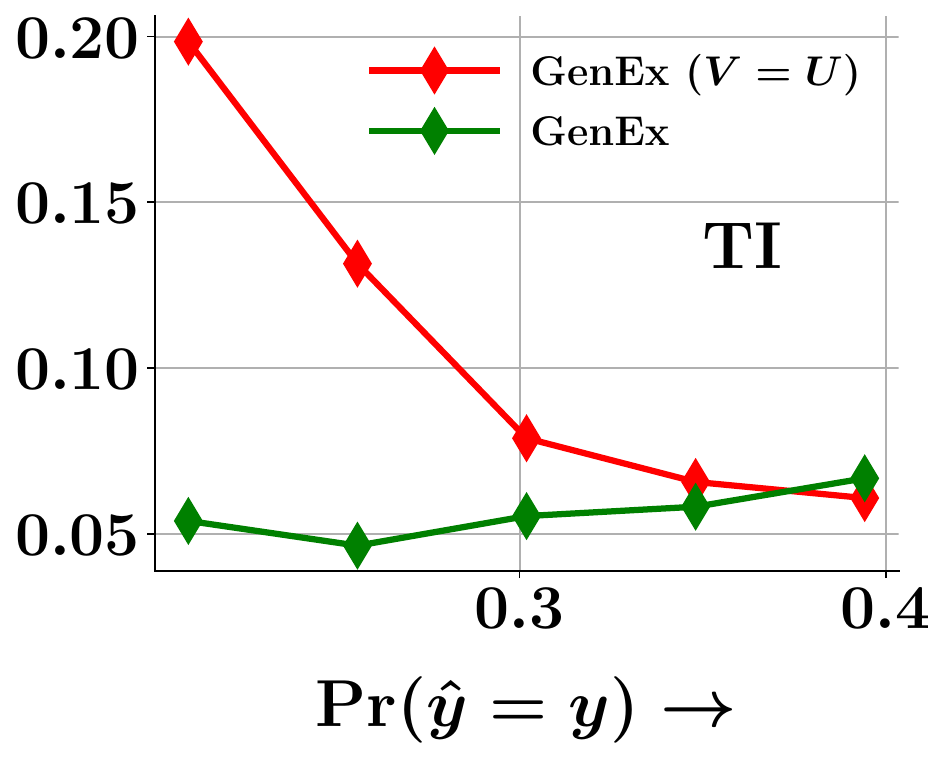}}  \\[-1ex]
    % \vspace{-1mm}
        \caption{\small Ablation study: Saved cost  vs. accuracy.}
        \label{fig:ablation2}
    \end{minipage}
    % \vspace{-1mm}
\end{figure}
\xhdr{Ablation study on the generator}
To evaluate the magnitude of cost saving that our generator provides, we compare \our\ against its two variants.  (I) \our\ ($V=\emptyset$): Here, all the features $\ucal{}$ of all the test instances are queried from the oracle.
(II) \our\ ($V=U$): Here, whenever an instance is qualified 
to have the features from the generator (the classifier confidence on the generated feature is high), all the features $\ucal{}$ are drawn from the generator. Figure \ref{fig:ablation1} shows the results in terms of accuracy vs. the budget of the oracle queries.  Figure \ref{fig:ablation2} shows the results in terms of the average fraction of the saved budget $\EE[|V|/|U|]$. Note that,  even in \our\ ($V=U$), not all instances result in high confidence on the generated features. In case of low confidence, we query the features from the oracle. 
We make the following observations:
\textbf{(1)} \our\ outperforms the other variants in most of the cases in terms of accuracy for a fixed budget (Fig.~\ref{fig:ablation1}).  At the places where we perform better, the gain is significant ($p<0.05$ for $\EE[U\cp V]\le 20\%$ for DP; $p<0.01$ for others).
\textbf{(2)} \our\ is able to save 3--5x cost at the same accuracy as compared to the \our\ ($\vcal{}=\ucal{}$) variant. 
\textbf{(3)} The fractional cost saved goes down as accuracy increases, since $\ucal{}$ increases, but $\lambda$ is fixed.

\begin{figure}[h!]
    \centering
\subfloat{\includegraphics[width=.4\textwidth]{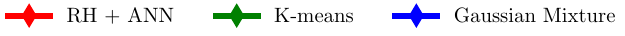}} \\[-0.1ex]
\subfloat{\includegraphics[width=.19\textwidth]{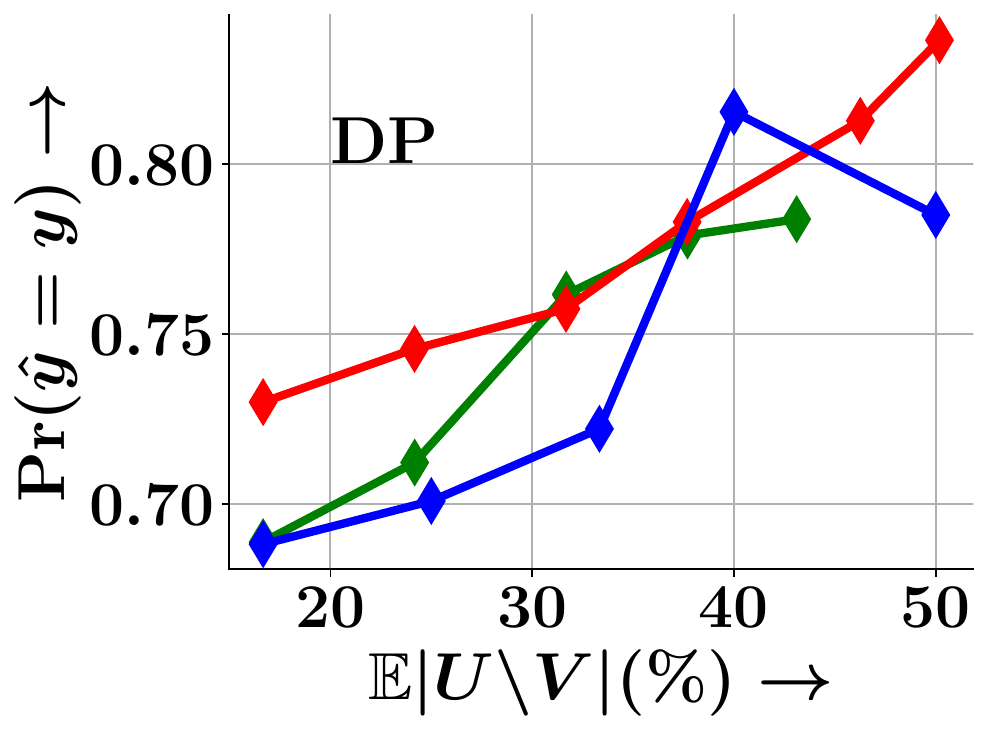}} \hspace{3mm}
\subfloat{\includegraphics[width=.18\textwidth]{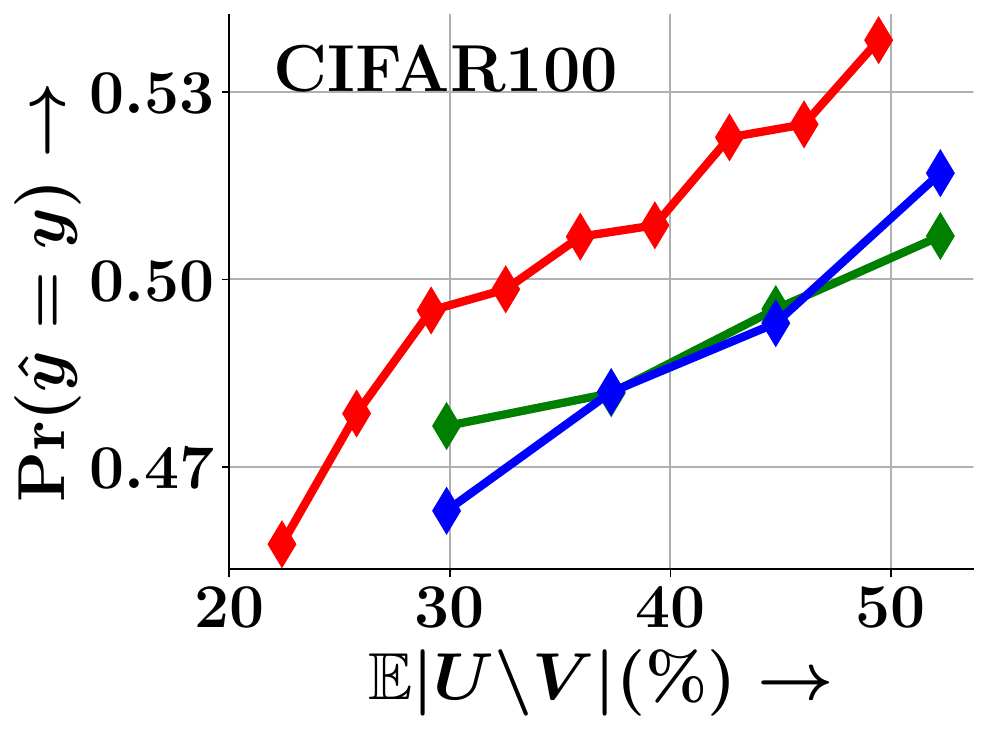}}\hspace{0.5mm}
% \vspace{-2mm}
\caption{\small RH vs. other clustering methods}
\label{fig:K-means}
\end{figure}

\xhdr{RH vs. other clustering methods}
Here, we compare random hyperplane (RH) guided clustering method with K-means and Gaussian mixture based clustering methods.
The results are summarized in Figure \ref{fig:K-means} for \dpr\ and \cifar\ datasets.  We observe that RH performs better for a wide range of oracle queries $\EE[|U\cp V|]$.
We note that the amount of bucket-skew, \ie, the ratio of the minimum and maximum size of buckets, is significantly better  for RH than K-means and GMM.  Specifically, for DP dataset, this ratio is $0.21$, $0.014$ and $0.003$ for RH, K-means and GMM, respectively. Thus, RH has a significantly better bucket balance than other methods. 
To further probe why the RH achieves a better bucket balance, we instrument the conicity of the features which is a measure of how the feature vectors are concentrated in a narrow cone centered at the origin~\cite{sharma2018towards}. This is defined as: 
$    \text{conicity}(D) = \frac{1}{|D|}\sum_{i\in D}\cos\left(\xb_i, {\sum_{j\in D} \xb_j}/{|D|} \right)$.
%
% we analyze how the feature vectors are concentrated in a narrow cone centered at the origin. We compute conicity \cite{sharma2018towards}, 
% $    \text{conicity}(D) = \frac{1}{|D|}\sum_{i\in D}\cos\left(\xb_i, {\sum_{j\in D} \xb_j}/{|D|} \right)$
% Higher the conicity, higher is the conentration of the feature vectors in a narrow cone.
% as defined in (\ref{eq:conicity}), to measure how the feature vectors are concentrated in a narrow cone centered at the origin. 
We observe a low conicity of $<0.2$, indicating a high spread of feature vectors. Since, on an average, the random hyperplanes cut the space uniformly across the origin, the observed feature vectors get equally distributed between 
different hyperplanes, leading to good bucket balance.
% The skew measured as the ratio of minimum and maximum size of buckets is $0.21$, $0.014$ and $0.003$ for LSH, K-means and GMM on the DP dataset. \textbf{(3)} 
Conversely, K-means and GMM maximize the "mean" of similarity, promoting a few highly similar points in one cluster and leaving moderately similar instances dispersed among different clusters. These methods tend to exhibit inter-cluster cosine similarities of at least 0.92, suggesting that these instances should ideally belong to the same cluster. However, due to their objective to maximize a single aggregate measure, K-means and GMM often group extremely similar items together while separating others.
\section{Conclusion}
% \vspace{-1mm}
We proposed \our, a model for acquiring subsets of features in a batch setting to maximize classification accuracy under a budget constraint. \our\ relies on a mixture of experts model with random hyperplane guided data partitioning and uses a generator to produce subsets of features at no additional query cost. We employ a greedy algorithm that takes the generated features into account and provides feature subsets for each data partition. We also introduce the notions of $(\mmin, \mmax)$-partial monotonicity and $(\gmin, \gmax)$-weak submodularity, and provide a theoretical foundation for our method. \our\ is superior to the baselines, outperforming them in accuracy at a fixed budget. We recognize that a limitation of our work is that the guarantee of the greedy algorithm holds under certain assumptions, which are an artifact of the complexity of the problem. 
Further work can be done incorporating explicit exploration-exploitation on the greedy strategy.

\bibliography{afa,refs}

\newpage
\appendix
\onecolumn
% \section{Appendix}

\begin{center}
    \Large { \bf Appendix \\
    \normalsize (Generator Assisted  Mixture of Experts   For Feature Acquisition in Batch)
    }
\end{center}

\section{Limitations}
\label{app:lim}
Limitations of our work are described as follows:
% It is possible to consider a setting, where multiple batches of features, each of variable length, must be acquired sequentially. This has practical application in the medical setting, in which a doctor may prescribe batches of tests multiple times before diagnosing a disease. We do not address this dimension of the feature selection problem. 

\textbf{(1)} The theoretical guarantees on \our\ hold under certain assumptions, which may not always be strictly true. However \our\ works very well in practice. 

\textbf{(2)} The greedy algorithm does not provide sufficient exploration in the space of features. This can possibly be remedied by an $\ee$-greedy algorithm, which provides an exploration-exploitation trade off.

\textbf{(3)} We did not assume an active setting in our setup during training. For example, if we only have the observed values and have no access to any unobserved values during training itself, then our algorithm cannot be directly applied. 

\textbf{(4)} Some features can have adverse instance specific cost. E.g., CT scan can be more harmful for some people than others. We did not model cost in an instance specific manner, in our work. It would be important to design instance specific cost per each new feature, \ie, $q(u \given \xb[\ocal{i}])$ where $u\in U$.

\section{Broader Impact}
\label{app:broader}
Our method can be used for feature acquisition in batch. Such a problem can be very useful in medical testing, where the goal is to suggest new tests to the patients.
A negative consequence can be querying for feature that may result in biased prediction. A feature acquisition algorithm can also query for sensitive features. Designing methods that can identify which feature is correlated with sensitive features, and preventing their acquisition can be very useful in practice.

Some states may have regulations about specific features. This include medical features, personal information, etc. In all such cases, it would be important to design a censored feature acquisition problem, which would query features that are informative, but neither violating any existing law and nor are correlated with any such features that are potentially sensitive.

\newpage
\section{Additional discussions on related work}
\label{app:related}

Apart from the works on sequential feature acquisition, discussed in Introduction, our work is related to theoretical works on online feature acquisition models, active learning and subset selection. In the following, we briefly review them.

\xhdr{Feature acquisition for online linear models}  A wide body of work~\cite{foster2016online,kale2017adaptive,ito2017efficient,murata2018sample} looks into theoretical problems related to online feature acquisition settings.
~\citet{foster2016online} provided an inefficient algorithm for online linear regression, which admits $\tilde{O}(\sqrt{T})$ regret. Moreover, they 
showed that there exists no polynomial time algorithm per iteration, that can achieve a regret of $O(T^{1-\delta})$ for $\delta>0$.  This result 
paved the way of finding new practical assumptions that can lead to design a tractable algorithm.
In this context,~\citet{kale2017adaptive} showed that the assumption of restricted isometry property (RIP) of the feature vectors 
can allow us to design efficient algorithm for the feature acquisition under an online linear regression setup.
~\citet{ito2017efficient} designed efficient algorithms under two closely connected assumptions with RIP, \viz,
linear independence of feature vectors and compatibility.~\citet{murata2018sample}
proposed stochastic gradient methods under linear regression setting and provided sample complexity under a restricted eigenvalue condition.

\xhdr{Active learning} The goal of active learning~\cite{active1,active2,active3,active4,active5,active6,active7} is to select a subset of unlabeled instances for labelling, so that model trained on those instances provide highest accuracy.~\citet{activeSurvey} provides a comprehensive survey.
In contrast, our goal is to query the values few feature entries, so that together with observed features, these newly acquired features provides us accurate predictions.
Hence, at a very high level, one can think of active learning as an instance of data subset selection, whereas our problem is feature subset selection.

\xhdr{Feature selection} Feature selection mechanisms have wide variety of applications in machine learning. These mechanisms involve observing the values of features before selecting the subset. Moreover, the subset is fixed for all instances. Among these ~\cite{khanna2017scalable,killamsetty2021grad,harshaw2019submodular,das2018approximate} use submodular optimization, which are based on greedy algorithms similar to ~\cite{elenberg2018restricted,mirzasoleiman2015lazier} . Mutual information measures~\cite{estevez2009normalized} and dependence maximisation~\cite{song2012feature} have been used to identify most relevant features. In the context of linear regression often LASSO is used to create sparsity in the weights, thus reducing the dependence on a subset of the features~\cite{hans2009bayesian} . In case of neural networks, feature selection has been done using stochastic gates as in~\cite{yamada2020feature} . Wrapper methods, which involve optimizing the model for each subset of features are computationally expensive in case of deep neural networks.
% \todo[inline]{Vedang, write in details about feature selection work}

\newpage
\section{Proofs of all technical results}
\label{app:proofs}

\subsection{Proof of Proposition~\ref{prop:0}}
% For any instance $i \in D$ the set of observed features to be $S^i$, the subset of the unobserved features which is directly queried from the data to be $T^i$ and the subset of unobserved features that are sampled from the generator to be $T_g^i$. Also, suppose the classifier is parameterized by $\theta$ and the generator by $\phi$. Then, our optimization problem is
% Our objective function for minimizing $U_b$ for fixed $\theta, \phi$ is as follows:
\begin{proof}
Consider the optimization problem~\eqref{eq:obj} for a given parameter set $\theta$ and $\phi$
\begin{align}
\min_{\set{U_i,V_i}: |U_i \cp V_i| \le q_{\max}}\sum_{i\in D_b}\EE_{\xg_i [\sub_i] \sim    p(\bullet \given     \xb_i[\obs_i  ])} \Big[ \ell\left(h\big(\xb_i [\obs_i \cup \unobs_i \cp\sub_i] \cup \xg_i [\sub_i]\big), y_i\right) \Big] \label{eq:mini}
\end{align}
Consider a special realization where for some instance $i \in Z$, $U_i = U_b, V_i = \emptyset$ and for $i\not\in Z$, we have $U_i=U_b, V_i = U_b$. They always satisfy $|U_i\cp V_i| \le q_{\max}$, as long as $|U_b| \le q_{\max}$. Thus, \emph{for any such $Z$}, the optimal value of the problem~\eqref{eq:mini} will be less than
\begin{align}
    \min_{U_b: |U_b| \le q_{\max}} 
  &  \sum_{i\in D_b\cp Z}\EE_{\xg_i [\sub_b] \sim    p(\bullet \given     \xb_i[\obs_i  ])} \Big[ \ell\left(h\big(\xb_i [\obs_i ] \cup \xg_i [U_b]\big), y_i\right) \Big]\nonumber\\
&  + \sum_{i\in  Z}  \ell\left(h\big(\xb_i [\obs_i \cup \unobs_b]\big), y_i\right) \label{eq:minj}
\end{align}
Now, we impose  probability $\Pr(i\in Z) = \Delta_i(U_b)$. Since the above quantity~\eqref{eq:minj} is larger than the optimal value of Eq.~\eqref{eq:mini} for any value of $Z$, the expected value of Eq.~\eqref{eq:minj} over $Z$ will be more than Eq.~\eqref{eq:mini}. Applying Jensen inequality ($\min (\bullet)$ is a concave function), we obtain that the expected value of Eq.~\eqref{eq:minj} is less than $\min_{U_b:|U_b| \le q_{\max}}\sum_{i\in D_b} F(h,p; U_b \given     \ocal{i})$.
\end{proof}

\subsection{Proof of Theorem~\ref{thm:mon}}
\begin{proof}
First we prove part (1), which is on the partial monotonicity of
$\gf$. We first evaluate $\gf(T) - \gf(S)$. To this end, we define: 
$\el_i(\phi_b , \theta _b , S)  =  \EE_{\xg \sim p_{\phi_b} (\bullet \given     \ocal{i})} \ell(h_{\theta_b }(\xb_i[\ocal{i}] \cup \xg_i[S]) ,y_i) $ and 
$\Lcal_i(\theta _b , S )  =   \ell(h_{\theta_b }(\xb_i[\ocal{i} \cup S]),y_i ) $ and 
\begin{align}
    \gf & (T) - \gf(S) \nonumber\\
    & = \sum_{i\in D_b}\left[(1-\Delta_i(T)) \cdot \el_i(\phi_b ^*(T) , \theta_b ^*(T), T) + \Delta_i(T) \cdot \Lcal_i (\theta _b ^*(T), T) \right]\nonumber \\
 &   \quad - \sum_{i\in D_b} \left[(1-\Delta_i(S)) \cdot \el_i(\phi_b ^*(S) , \theta_b ^*(S), S) + \Delta_i(S) \cdot \Lcal_i (\theta _b ^*(S), S) \right]  
 \end{align}
Adding and subtracting the term: $\sum_{i\in D_b}\left[(1-\Delta_i(T)) \cdot \el_i(\phi_b ^*(S) , \theta_b ^*(S), S) + \Delta_i(T) \cdot \Lcal_i (\theta _b ^*(S), S) \right]$
we obtain the following:
 \begin{align}
 \gf & (T) - \gf(S) \nonumber\\
& = \sum_{i\in D_b}\left[(1-\Delta_i(T)) \cdot \el_i(\phi_b ^*(T) , \theta_b ^*(T), T) + \Delta_i(T) \cdot \Lcal_i (\theta _b ^*(T), T) \right]\nonumber \\
&\quad - \sum_{i\in D_b}\left[(1-\Delta_i(T)) \cdot \el_i(\phi_b ^*(S) , \theta_b ^*(S), S) + \Delta_i(T) \cdot \Lcal_i (\theta _b ^*(S), S) \right]\nonumber \\
&\quad + \sum_{i\in D_b}\left[(1-\Delta_i(T)) \cdot \el_i(\phi_b ^*(S) , \theta_b ^*(S), S) + \Delta_i(T) \cdot \Lcal_i (\theta _b ^*(S), S) \right]\nonumber \\
 &   \quad - \sum_{i\in D_b} \left[(1-\Delta_i(S)) \cdot \el_i(\phi_b ^*(S) , \theta_b ^*(S), S) + \Delta_i(S) \cdot \Lcal_i (\theta _b ^*(S), S) \right] \label{eq:gf0}
\end{align}
Now, $\phi_b ^* (T)$ and $\theta_b ^*(T)$ are the optimal parameters of $T \supset S$. The support of $\theta_b ^*(T)$
is subset of the support of $\theta_b ^* (S)$. 
Hence, 
\begin{align}
&    \sum_{i\in D_b}\left[(1-\Delta_i(T)) \cdot \el_i(\phi_b ^*(T) , \theta_b ^*(T), T) + \Delta_i(T) \cdot \Lcal_i (\theta _b ^*(T), T) \right] \nonumber\\
&\qquad  \le  \sum_{i\in D_b} \left[(1-\Delta_i(T)) \cdot \el_i(\phi_b ^*(S) , \theta_b ^*(S), S) + \Delta_i(T) \cdot \Lcal_i (\theta _b ^*(S), S) \right]
\end{align}
because: if this is not true, then we can set $\theta_b$ such that $\theta_b [S] = \theta_b ^*(S)$ and $\theta_b [T\cp S] = 0$, which would be a minimizer of $  \sum_{i\in D_b}\left[(1-\Delta_i(T)) \cdot \el_i(\phi_b ^*(T) , \theta_b, T) + \Delta_i(T) \cdot \Lcal_i (\theta _b, T) \right]$.

Putting this relation into the Eq.~\eqref{eq:gf0}, we have:
  \begin{align}
   \gf(T) - \gf(S) & \le    \sum_{i\in D_b}\left[(1-\Delta_i(T)) \cdot \el_i(\phi_b ^*(S) , \theta_b ^*(S), S) + \Delta_i(T) \cdot \Lcal_i (\theta _b ^*(S), S) \right] \nonumber \\
& \qquad  - \sum_{i\in D_b} \left[(1-\Delta_i(S)) \cdot \el_i(\phi_b ^*(S) , \theta_b ^*(S), S) + \Delta_i(S) \cdot \Lcal_i (\theta _b ^*(S), S) \right] \\
& \qquad \le |\Delta_i(T) - \Delta_i(S)| |\el_i(\phi_b ^*(S) , \theta_b ^*(S), S) - \Lcal_i(\theta_b ^*(S), S) |\\
& \qquad \le |D_b| L_{x}\ee_x \ee_\Delta 
\end{align}
Furthermore $\gf(S) > |D_b|\ell_{\min}$ gives us that $\gf(T)/\gf(S) \le 1+\frac{L_x \ee_x \ee_\Delta }{\ell_{\min}}  $. Next we compute lower bound on $\gf(T)/\gf(S)$. To do so we note that:
\begin{align}
    \gf&(S) - \gf(T)\nonumber\\
    & = \sum_{i\in D_b}\left[(1-\Delta_i(S)) \cdot \el_i(\phi_b ^*(S) , \theta_b ^*(S), S) + \Delta_i(S) \cdot \Lcal_i (\theta _b ^*(S), S) \right]\nonumber \\
 &   \quad - \sum_{i\in D_b} \left[(1-\Delta_i(T)) \cdot \el_i(\phi_b ^*(T) , \theta_b ^*(T), T) + \Delta_i(T) \cdot \Lcal_i (\theta _b ^*(T), T) \right] \nonumber\\
 & \quad = \sum_{i\in D_b}(\Delta_i(S) -\Delta_i(T))\cdot \Lcal_i (\theta _b ^*(S), S) + \Delta_i(T)\cdot [ \el_i(\phi_b ^*(T) , \theta_b ^*(T), T)  - \Lcal_i( \theta_b ^*(T), T)] \nonumber \\
 &  \quad  + \sum_{i\in D_b} \Delta_i (T) \cdot \Lcal_i (\theta _b ^*(S), S) -\el_i(\phi_b ^*(T) , \theta_b ^*(T), T) +[1-\Delta_i(S)] \cdot \el_i(\phi_b ^*(S) , \theta_b ^*(S), S) \nonumber \\
 & \le |D_b|[ \ell_{\max} \ee_\Delta + \Delta_{\max} \ee_x + 2\ell_{\max} (1+\Delta_{\max})]
 \end{align}
 Thus, $\displaystyle  \frac{\gf(T)}{\gf(S)} \ge
 \left[1+   \frac{\ell_{\max} \ee_{\Delta}}{\ell_{\min}} 
  + \frac{L_x \Delta_{\max} \ee_x}{\ell_{\min}} + \frac{2\ell_{\max} (1+\Delta_{\max})}{\ell_{\min}}\right]^{-1}$
  
  Next, we prove part (2). We note that for any two sets $S,S'$ (not necessarily subsets of each other) 
\begin{align}
\gl(S)-\gl(S') \le  |D_b| \frac{\partial \ell(h(\xb),y)}{\partial \xb} \bigg|_{\max} \ee_x
 \end{align}
 This gives us the required bound.
\end{proof}
% \newpage
 \subsection{Proof of Theorem~\ref{thm:sub}}
 
% \begin{numtheorem}{~\ref{thm:sub}} Let $\ell(h_{\theta}(\xb),y)$  be strictly convex in $\theta$.  Moreover,  assume that 
% $\nabla_{\theta}\ell(h_{\theta}(\xb),y) \le \nabla_{\max}$ 
% and $\text{Eigenvalues}\set{\nabla^2 _{\theta} \ell(h_{\theta}(\xb[S]),y)} \in [\zeta_{\min}, \zeta_{\max}]$ for all $S$; the parameters of the generator 
% $\phi$ is bounded, \ie, $||\phi|| \le \phi_{\max}$; $\Delta_{\max} > \max\set{0,1-{\nabla^2 _{\max}}/{4\zeta_{\max} \phi_{\max} 
% L_{\phi}}} $ and $\ee_{\Delta} \ee_x < {\nabla^2 _{\max}}/{2\zeta_{\max}L_{x}} - {(1-\Delta_{\max}) \phi_{\max} L_{\phi} }/{L_{x}} $ 
% Then, we have the following result:
% The set function $\gf$  is $(\gmin_F,\gmax_F)$-weakly submodular 
% with
% \begin{align}
% \hspace{-2mm}\gmin_F \ge  \frac{0.5\nabla_{\max} ^2/ \zeta_{\max}-2 L_{\phi} \phi_{\max} - \ee_x \ee_{\Delta} L_x}{0.5\nabla_{\max} ^2/\zeta_{\min} +2 L_{\phi} \phi_{\max} + \ee_x \ee_{\Delta} L_x }, \
% \gmax_F \le \frac{2 L_{\phi} \phi_{\max} + \ee_x \ee_{\Delta} L_x}{0.5\nabla_{\max} ^2/\zeta_{\max} -2 L_{\phi} \phi_{\max} - \ee_x \ee_{\Delta} L_x }
% \end{align}
% \end{numtheorem}

\begin{proof}
Following the proof of Theorem~\ref{thm:mon},
let us define
$\el_i(\phi_b , \theta _b , S)  =  \EE_{\xg \sim p_{\phi_b} (\bullet \given     \ocal{i})} \ell(h_{\theta_b }(\xb_i[\ocal{i}] \cup \xg_i[S]) ,y_i) $ and 
$\Lcal_i(\theta _b , S )  =   \ell(h_{\theta_b }(\xb_i[\ocal{i} \cup S]),y_i ) $. Additionally, we  
% We would like to highlight that $L_\phi$ is the Lipschitz constant of $F$ with respect to $\phi$ and $L_x$ is the Lipschitz constant of $\ell(h_{\theta_b}(\xb),y)$ with respect to $\xb$.
define:
\begin{align}
    A(\phi,  \theta, T \given \set{\Delta_i}) =   \sum_{i\in D_b}\left[(1-\Delta_i) \cdot \el_i(\phi , \theta,T) + \Delta_i \cdot \Lcal_i (\theta,T) \right]
\end{align}
Given $\ell(h_\theta(\xb),y)$ is convex in $\theta$, 
$A (\phi,\theta,T)$ is convex $\theta$ and thus the function  $-A(\phi,\theta, T)$ is concave in $\theta$.
Following the results of~\citet[Proof of Theorem 5]{elenberg2018restricted}, we have the following relationship:
\begin{align}
& \hspace{-6mm} A( \phi,  \theta^* _b (T), T \cup S) \given \set{\Delta_i})- A(\phi,  \theta^* _b (T\cup S), T\cup S \given \set{\Delta_i})     \in \left[\frac{|D_b|\nabla_{\max} ^2}{2\zeta_{\max}},\frac{|D_b|\nabla_{\max} ^2}{2\zeta_{\min}}\right] \label{eq:E0} \\
& \hspace{-6mm} \sum_{s\in S}\bigg( A(\phi,  \theta^* _b (T), T \cup s \given \set{\Delta_i})- A(\phi,  \theta^* _b (T\cup s), T\cup s \given \set{\Delta_i})  \bigg)\in \left[\frac{|D_b|\nabla_{\max} ^2}{2\zeta_{\max}},\frac{|D_b|\nabla_{\max} ^2}{2\zeta_{\min}}\right] \label{eq:E1}
\end{align}
~\citet{elenberg2018restricted} proved the results using Taylor series expansion upto order two. Note that, $\phi_b ^* (T\cup S)$ remains unchanged across  in  the arguments of the first and second terms of LHSs  in Eqs.~\eqref{eq:E0} and~\eqref{eq:E1} and the only change of variable is $\theta _b ^* (T) \to \theta _b ^* (T\cup S)$.
 
To compute $\gmin_F$ and $  \gmax_F$ for $\gf$, we need to bound $\displaystyle \sum_{s\in S}\frac{\gf(s \given     T)}{|\gf(S \given     T)|}$.  For any set $S\neq \emptyset$,  we have the following marginal gain:
\begin{align}
 &  \gf(S \given T)\nn\\
 &   =  \gf   (T\cup S) - \gf(T)  \nonumber\\
    & = {\sum_{i\in D_b}\left[(1-\Delta_i(T \cup S)) \cdot \el_i(\phi_b ^*(T \cup S) , \theta_b ^*(T \cup S),  T\cup S) + \Delta_i(T \cup S) \cdot \Lcal_i (\theta _b ^*(T \cup S),  T\cup S) \right]} \nonumber \\
 &   \qquad - \sum_{i\in D_b} \left[(1-\Delta_i(T)) \cdot \el_i(\phi_b ^*(T) , \theta_b ^*(T), T) + \Delta_i(T) \cdot \Lcal_i (\theta _b ^*(T), T) \right]   
 \end{align}
Adding and subtracting   $ \sum_{i\in D_b}\left[(1-\Delta_i(T \cup S)) \cdot \el_i(\phi_b ^*(T) , \theta_b ^*(T), T) + \Delta_i(T \cup S) \cdot \Lcal_i (\theta _b ^*(T), T) \right]$
to the above, we obtain
 \begin{align}
  \gf   (T\cup S) & - \gf( T) \nonumber\\
& =  \underbrace{\sum_{i\in D_b}\left[(1-\Delta_i(T \cup S)) \cdot \el_i(\phi_b ^*(T \cup S) , \theta_b ^*(T \cup S), T\cup S) + \Delta_i(T \cup S) \cdot \Lcal_i (\theta _b ^*(T \cup S),  T\cup S) \right]}_{ \displaystyle A(\phi_b ^* (T\cup S), \theta_b ^* (T\cup S), T\cup S) \given \set{\Delta_i (T\cup S)} }\nonumber \\[3ex]
 &   \qquad -  \underbrace{\sum_{i\in D_b}\left[(1-\Delta_i(T \cup S)) \cdot \el_i(\phi_b ^*(T) , \theta_b ^*(T), T) + \Delta_i(T \cup S) \cdot \Lcal_i (\theta _b ^*(T), T) \right]}_{\displaystyle A(\phi_b ^* (T), \theta_b ^* (T), T\given \set{\Delta_i (T\cup S)})}\nonumber  \\[3ex]
 & \quad +   \sum_{i\in D_b}\left[(1-\Delta_i(T \cup S)) \cdot \el_i(\phi_b ^*(T) , \theta_b ^*(T), T) + \Delta_i(T \cup S) \cdot \Lcal_i (\theta _b ^*(T), T) \right]\nonumber \\
  &   \qquad - \sum_{i\in D_b} \left[(1-\Delta_i(T)) \cdot \el_i(\phi_b ^*(T) , \theta_b ^*(T), T) + \Delta_i(T) \cdot \Lcal_i (\theta _b ^*(T), T) \right]   \nonumber \\ 
  &  = A(\phi^* _b (T\cup S),  \theta^* _b (T\cup S), T\cup S \given \set{\Delta_i (T\cup S)}) - A(\phi^* _b (T),  \theta^* _b (T), T \given \set{\Delta_i (T\cup S)})  \nn \\
 & \qquad   - \sum_{i\in D_b}(\Delta_i(T\cup S) - \Delta_i (S) ) ( \el_i(\phi_b ^*(T) , \theta_b ^*(T), T) - \Lcal_i (\theta _b ^*(T), T) ) \label{eq:int0} 
 \end{align}
Now, since the support of 
$ \theta^* _b (T)$ is only limited to the entries $T$, we have 
\begin{align}
A(\phi^* _b (T\cup S),  \theta^* _b (T), T\cup S \given \set{\Delta_i (T\cup S)})  = A(\phi^* _b (T\cup S),  \theta^* _b (T), T \given \sde) 
\end{align}
Thus, we add $A(\phi^* _b (T\cup S),  \theta^* _b (T), T\cup S \given \sde) $ and subtract  $A(\phi^* _b (T\cup S),  \theta^* _b (T), T \sde) $
to Eq.~\eqref{eq:int0} and have:
 \begin{align}
   \gf  (T\cup S) - \gf(T) &  =  A(\phi^* _b (T\cup S),  \theta^* _b (T\cup S), T\cup S \given \sde) - A(\phi^* _b (T\cup S),  \theta^* _b (T), T\cup S \given \sde)\nonumber \\
&\quad  +  A(\phi^* _b (T\cup S),  \theta^* _b (T), T \given \sde)  - A(\phi^* _b (T),  \theta^* _b (T), T \given \sde) \nonumber \\
  &\quad - \sum_{i\in D_b}(\Delta_i(T\cup S) - \Delta_i (S) ) ( \el_i(\phi_b ^*(T) , \theta_b ^*(T), T) - \Lcal_i (\theta _b ^*(T), T) ) \label{eq:EE}
\end{align}
% Now, we will compute upper and lower bounds of the following quantities:
% \begin{enumerate}
%     \item $A(\phi^* _b (T\cup S),  \theta^* _b (T\cup S), T\cup S) - A(\phi^* _b (T\cup S),  \theta^* _b (T), T \cup S)$
%     \item $ A(\phi^* _b (T\cup S),  \theta^* _b (T), T)  - A(\phi^* _b (T),  \theta^* _b (T), T)$ 
%     \item $ \sum_{i\in D_b}(\Delta_i(T\cup S) - \Delta_i (S) ) ( \el_i(\phi_b ^*(T) , \theta_b ^*(T), T) - \Lcal_i (\theta _b ^*(T), T) )$
% \end{enumerate}
%
% \xhdr{Upper and lower bound on $A(\phi^* _b (T\cup S),  \theta^* _b (T\cup S), T\cup S) - A(\phi^* _b (T\cup S),  \theta^* _b (T), T \cup S)$}
First, we bound the second term and the third terms of RHS here, which will be used in bounding both numerator and denominator of $\frac{\sum_{s\in S} \gf(s\in T)}{|\gf(S\given T)|}$. Using Lipshitz continuity, we have:
\begin{align}
 &A(\phi^* _b (T\cup S),  \theta^* _b (T), T \given \sde)  - A(\phi^* _b (T),  \theta^* _b (T), T \given \sde) \nonumber\\
 & \in [-|D_b| L_{\phi} ||\phi^* _b (T\cup S) -\phi^* _b (T)||,|D_b| L_{\phi} ||\phi^* _b (T\cup S) -\phi^* _b (T)||] \nonumber \\
& \in [-2 |D_b|\cdot L_{\phi} \cdot \phi_{\max},2 |D_b| L_{\phi} \phi_{\max}] \label{eq:b1}
\end{align}
where $L_{\phi}$ is the Lipschitz constant of $F$ with respect to $\phi$ and $||\phi|| < \phi_{\max}$. Finally, to bound the third term,   we have:
 \begin{align}
      \sum_{i\in D_b}(\Delta_i(T\cup S) - \Delta_i (S) ) ( \el_i(\phi_b ^*(T) , \theta_b ^*(T), T) - \Lcal_i (\theta _b ^*(T), T) ) \in [-|D_b| \ee_x \ee_\Delta L_x,|D_b| \ee_x \ee_\Delta L_x]\nonumber\\[-2ex]\label{eq:b2}
 \end{align}
 Here, $L_x$ is the Lipschitz constant of $\ell(h_{\theta_b}(\xb),y)$ with respect to $\xb$.

\xhdr{Bounds on $\sum_{s\in S}G(s\given T)$:} Eq.~\eqref{eq:EE} is valid for any $S$. If we set $S = \set{s}$ in Eq.~\eqref{eq:EE} and then take a sum over all $s$, then we have:
 \begin{align}
  \sum_{s \in S} [\gf (T\cup s) - \gf(T)] &  = \sum_{s\in S} A(\phi^* _b (T\cup s),  \theta^* _b (T\cup s), T\cup s \given \sdes) - A(\phi^* _b (T\cup s),  \theta^* _b (T), T\cup s \given \sdes)\nonumber \\
&\quad  + \sum_{s\in S} A(\phi^* _b (T\cup s),  \theta^* _b (T), T \given \sdes)  - A(\phi^* _b (T),  \theta^* _b (T), T \given \sdes) \nonumber \\
  &\quad - \sum_{s\in S} \sum_{i\in D_b}(\Delta_i(T\cup s) - \Delta_i (S) ) ( \el_i(\phi_b ^*(T) , \theta_b ^*(T), T) - \Lcal_i (\theta _b ^*(T), T) ) \label{eq:BB-1}
\end{align}
From Eq.~\eqref{eq:E1}  we note that, 
\begin{align}
-\frac{|D_b|\nabla_{\max} ^2}{2\zeta_{\min}} \le     A(\phi^* _b (T\cup s),  \theta^* _b (T\cup s), T\cup s \given \sdes) - A(\phi^* _b (T\cup s),  \theta^* _b (T), T \cup s \given \sdes) \le -\frac{|D_b|\nabla_{\max} ^2}{2\zeta_{\max}} \label{eq:A-A-0}
\end{align}
% Using the three bounds given in Eqs.~\eqref{eq:A-A-0}, ~\eqref{eq:b1} and~\eqref{eq:b2} in Eq.~\eqref{eq:BB-1}, we have:
% \begin{align}
%       \sum_{s \in S} [\gf (T\cup s) - \gf(T)] &  = \sum_{s\in S} A(\phi^* _b (T\cup s),  \theta^* _b (T\cup s), T\cup s \given \sdes) - A(\phi^* _b (T\cup s),  \theta^* _b (T), T\cup s \given \sdes)\nonumber \\
% &\quad  + \sum_{s\in S} A(\phi^* _b (T\cup s),  \theta^* _b (T), T \given \sdes)  - A(\phi^* _b (T),  \theta^* _b (T), T \given \sdes) \nonumber \\
%   &\quad - \sum_{s\in S} \sum_{i\in D_b}(\Delta_i(T\cup s) - \Delta_i (S) ) ( \el_i(\phi_b ^*(T) , \theta_b ^*(T), T) - \Lcal_i (\theta _b ^*(T), T) ) \label{eq:BB-1}
% \end{align}
% Moreover, using Lipshitz continuity, we have:
% \begin{align}
%  &A(\phi^* _b (T\cup S),  \theta^* _b (T), T)  - A(\phi^* _b (T),  \theta^* _b (T), T) \nonumber\\
%  & \in [-|D_b| L_{\phi} ||\phi^* _b (T\cup S) -\phi^* _b (T)||,|D_b| L_{\phi} ||\phi^* _b (T\cup S) -\phi^* _b (T)||] \nonumber \\
% & \in [-2 |D_b|\cdot L_{\phi} \cdot \phi_{\max},2 |D_b| L_{\phi} \phi_{\max}] \label{eq:b1}
% \end{align}
% where $L_{\phi}$ is the Lipschitz constant of $F$ with respect to $\phi$. Finally, to bound the third term  we have:
%  \begin{align}
%       \sum_{i\in D_b}(\Delta_i(T\cup S) - \Delta_i (S) ) ( \el_i(\phi_b ^*(T) , \theta_b ^*(T), T) - \Lcal_i (\theta _b ^*(T), T) ) \in [-|D_b| \ee_x \ee_\Delta L_x,|D_b| \ee_x \ee_\Delta L_x]\nonumber\\[-2ex]\label{eq:b2}
%  \end{align}
%  Here, $L_x$ is the Lipschitz constant of $\ell(h_{\theta_b}(\xb),y)$ with respect to $\xb$.
 Using \emph{the upper bounds} in Eqs.~\eqref{eq:A-A-0},~\eqref{eq:b1} and~\eqref{eq:b2} in Eq.~\eqref{eq:BB-1}, we have the upper bound on $\sum_{s\in S} \gf(s\given T) = \sum_{s \in S} [\gf (T\cup s) - \gf(T)]$ as:
 \begin{align}
 \sum_{s\in S} G(s\given     T) \le  -\frac{|D_b|\nabla_{\max} ^2}{2\zeta_{\max}}+2n|D_b| L_\phi \phi_{\max} + n|D_b| \ee_{\Delta} \ee_x L_x. 
 \end{align}
 Using \emph{the lower bounds} in Eqs.~\eqref{eq:A-A-0},~\eqref{eq:b1} and~\eqref{eq:b2} in Eq.~\eqref{eq:BB-1}, we have the upper bound on $\sum_{s\in S} \gf(s\given T)$ as:
 % Similarly applying the bounds derived from~\citet{elenberg2018restricted} presented in Eq.~\eqref{eq:E0} and the  lower bounds from ~\eqref{eq:b1} and~\eqref{eq:b2} we obtain the lower bound on $\sum_{s\in S} G(s\given     T)$, \ie, 
\begin{align} 
\sum_{s\in S} G(s\given     T) \ge  -\frac{|D_b|\nabla_{\max} ^2}{2\zeta_{\min}}-2n|D_b| L_\phi \phi_{\max} - n|D_b| \ee_{\Delta} \ee_x L_x .
\end{align}
\xhdr{Bounds on $\sum_{s\in S}G(s\given T)$:}  
First, by applying triangle inequalities $|a|-|b|\le |a+b| \le |a|+|b|$, we have the upper bound:
\begin{align}
     \left| \gf(S\given     T) \right|
   & \le  \left|A(\phi^* _b (T\cup S),  \theta^* _b (T\cup S), T\cup S) - A(\phi^* _b (T\cup S),  \theta^* _b (T), T\cup S)\right|\nonumber \\
&\quad  + \left| A(\phi^* _b (T\cup S),  \theta^* _b (T), T)  - A(\phi^* _b (T),  \theta^* _b (T), T)\right|\nonumber \\
  &\quad +\left| \sum_{i\in D_b}(\Delta_i(T\cup S) - \Delta_i (S) ) ( \el_i(\phi_b ^*(T) , \theta_b ^*(T), T) - \Lcal_i (\theta _b ^*(T), T) )\right| \nn\\
&\qquad \le \frac{|D_b|\nabla_{\max} ^2}{2\zeta_{\min}}+2|D_b| L_\phi \phi_{\max} + |D_b| \ee_{\Delta} \ee_x L_x 
\label{eq:EE3}
\end{align}
The last inequality is due to the bounds in Eqs.~\eqref{eq:E0}, ~\eqref{eq:b1} and~\eqref{eq:b2}. Similarly, we obtain the lower bound on  $   \left| \gf(S\given     T) \right|$ as follows:
\begin{align}
     \left| \gf(S\given     T) \right|
   & \ge  \left|A(\phi^* _b (T\cup S),  \theta^* _b (T\cup S), T\cup S) - A(\phi^* _b (T\cup S),  \theta^* _b (T), T\cup S)\right|\nonumber \\
&\quad  - \left| A(\phi^* _b (T\cup S),  \theta^* _b (T), T)  - A(\phi^* _b (T),  \theta^* _b (T), T)\right|\nonumber \\
  &\quad -\left| \sum_{i\in D_b}(\Delta_i(T\cup S) - \Delta_i (S) ) ( \el_i(\phi_b ^*(T) , \theta_b ^*(T), T) - \Lcal_i (\theta _b ^*(T), T) )\right| \nn\\
&\qquad \le \frac{|D_b|\nabla_{\max} ^2}{2\zeta_{\max}}-2|D_b| L_\phi \phi_{\max} - |D_b| \ee_{\Delta} \ee_x L_x 
\label{eq:EE4}
\end{align}

Thus, finally, we have:
\begin{align}
 \overline{\gamma}_F \le \max\left\{  \frac{   -\frac{|D_b|\nabla_{\max} ^2}{2\zeta_{\max}}+2n|D_b| L_\phi \phi_{\max} + n|D_b| \ee_{\Delta} \ee_x L_x}{\frac{|D_b|\nabla_{\max} ^2}{2\zeta_{\min}}+2|D_b| L_\phi \phi_{\max} + |D_b| \ee_{\Delta} \ee_x L_x  }, 
 \frac{   -\frac{|D_b|\nabla_{\max} ^2}{2\zeta_{\max}}+2n|D_b| L_\phi \phi_{\max} + n|D_b| \ee_{\Delta} \ee_x L_x}{\frac{|D_b|\nabla_{\max} ^2}{2\zeta_{\max}}-2|D_b| L_\phi \phi_{\max} - |D_b| \ee_{\Delta} \ee_x L_x  }
 \right\}. 
\end{align}
\begin{align}
 \gmin_F \ge   
 \frac{   -\frac{|D_b|\nabla_{\max} ^2}{2\zeta_{\min}}-2n|D_b| L_\phi \phi_{\max} - n|D_b| \ee_{\Delta} \ee_x L_x}{\frac{|D_b|\nabla_{\max} ^2}{2\zeta_{\max}}-2|D_b| L_\phi \phi_{\max} - |D_b| \ee_{\Delta} \ee_x L_x  }
\end{align}
\end{proof}

 \subsection{Proof of Theorem~\ref{thm:greedy}}
 
% \begin{numtheorem}{~\ref{thm:greedy}}For bucket $b$, $\gf$ is a $(\mmin_F,\mmax_F)$-partially monotone, $(\gmin_F,\gmax_F)$-weakly submodular set function. Suppose $ \ucalx{T}$ is
% the output of $\textsc{GreedyForU}$. Then, if $OPT=\argmin_{U_b} \gf(U_b), |U_b| \le {q_{\max}}$, we have $\gf( \ucalx{T}) \le m_F \gf(OPT)-\left(1-\gamma_F/{q_{\max}} \right)^{q_{\max}} (m_F \gf(OPT)-\gf(\emptyset))$
% where $m_F=\max\left(\mmax_F, 2\mmax_F/\mmin_F \right)$ and $\gamma_F = \max(\gmax_F,-\gmin_F)$.
% \end{numtheorem}

\begin{proof}
From $(\mmin_F,\mmax_F)$-partially monotonicity for any $S,T$ where $S \subset T$, we have
\begin{align}
    \mmin_F &\le \frac{\gf(T)}{\gf(S)} \le \mmax_F
\end{align}
and from $(\gmin_F,\gmax_F)$-weakly submodularity, for any $S,T$, where $S \cap T = \emptyset$, we have
\begin{align}
    -\gamma_F &\le \frac{\sum_{u \in T} \gf(u\given    S)}{|\gf(T|S)|} \le \gamma_F
\label{eq:gamma_submod}
\end{align}
where $\gamma_F = \max(-\gmin_F, \gmax_F)$.

Suppose our greedy algorithm produces a sequence of iterates $\ucalx{0}, \ucalx{1} \cdots \ucalx{T}$, where $\ucalx{0} = \emptyset$ and $|\ucalx{T}| = \ucal{b}^*\le q_{\max}$. We have
\begin{align}
    \gf(\ucalx{i}) - \gf(\ucalx{i-1}) &= \min_{u \notin \ucalx{i-1}} \gf(u\given     \ucalx{i-1})\\
    \text{where } \min_{u \notin \ucalx{i-1}} \gf(u\given     \ucalx{i-1}) &\le 0 ~\quad \text{for all } i \in \{1,2 \cdots |\ucalx{T}|\}
\end{align}
Let $OPT=\argmin_{U_b} \gf(U_b), |U_b| \le {q_{\max}}$. Then, since $\min_{u \notin \ucalx{i-1}} \gf(u\given     \ucalx{i-1}) \le 0$ and $|OPT \cp \ucalx{i-1}| \le q_{\max}$
\begin{align}
    \gf(\ucalx{i}) - \gf(\ucalx{i-1}) &\le \frac{|OPT\cp \ucalx{i-1}|}{q_{\max}} \min_{u \notin \ucalx{i-1}} \gf(u\given     \ucalx{i-1})\\
                                    &\le \frac{|OPT\cp \ucalx{i-1}|}{q_{\max}} \min_{u \notin OPT \cp \ucalx{i-1}} \gf(u\given     \ucalx{i-1})\\
                                    &\le \frac{ \sum_{u\in OPT\cp \ucalx{i-1}} \gf(u\given     \ucalx{i-1}) }{q_{\max}}
    \label{eq:iter}
\end{align}
We now attempt to get bounds on the numerator of the RHS. Consider two cases

\emph{Case 1:} $\gf(OPT\cp \ucalx{i-1}) \ge 0$, then by Eq.~ \eqref{eq:gamma_submod}, with $S=\ucalx{i-1}$ and $T = OPT\cp \ucalx{i-1}$, we have
\begin{align}
    \sum_{u\in OPT\cp \ucalx{i-1}} \gf(u\given    \ucalx{i-1}) &\le \gamma_F[\gf(OPT \cup \ucalx{i-1})-\gf(\ucalx{i-1})] \\
                                &\le \gamma_F[\mmax_F\cdot \gf(OPT)-\gf(\ucalx{i-1})]
\end{align}

\emph{Case 2:} $\gf(OPT\cp \ucalx{i-1}) < 0$, then by Eq.~\eqref{eq:gamma_submod}, with $S=\ucalx{i-1}$ and $T = OPT\cp \ucalx{i-1}$, and noting that $\gf(S)\ge 0\ \forall\,  S$, we have
\begin{align}
    \sum_{u\in OPT\cp \ucalx{i-1}} \gf(u\given    \ucalx{i-1}) &\le \gamma_F[\gf(\ucalx{i-1}) - \gf(OPT \cup \ucalx{i-1})] \\
                                                    &\le \gamma_F \gf(\ucalx{i-1})\\
                                                    &= \gamma_F [2\gf(\ucalx{i-1}) - \gf(\ucalx{i-1})]\\
                                                    &\le \gamma_F \left[ \frac{2}{\mmin_F} \gf(OPT \cup \ucalx{i-1}) - \gf(\ucalx{i-1})\right]\\
                                                    &\le \gamma_F \left[ \frac{2\mmax_F}{\mmin_F} \gf(OPT) - \gf(\ucalx{i-1})\right]
\end{align}
Let $m_F=\max\left(\mmax_F, 2\mmax_F/\mmin_F \right)$, then,
\begin{align}
    \sum_{u\in OPT\cp \ucalx{i-1}} \gf(u\given    \ucalx{i-1}) &\le \gamma_F[m_F \gf(OPT) - \gf(\ucalx{i-1})]
    \label{eq:m_f}
\end{align}
Combining Eqs.~\eqref{eq:iter} and \eqref{eq:m_f}, we get
\begin{align}
    \gf(\ucalx{i}) - \gf(\ucalx{i-1}) &\le \frac{\gamma_F}{q_{\max}} \left[ m_F \gf(OPT) - \gf(\ucalx{i-1})\right]\\
    \implies  m_F \gf(OPT) - \gf(\ucalx{i}) &\ge \left( 1 - \frac{\gamma_F}{q_{\max}} \right) \bigg[m_F \gf(OPT) - \gf(\ucalx{i-1})\bigg]
\end{align}
If we use this relation recursively for $  i \in \{1,2 \cdots | \ucalx{T}|\}$, we get
\begin{align}
    m_F \gf(OPT) - \gf( \ucalx{T}) &\ge \left( 1 - \frac{\gamma_F}{q_{\max}} \right)^{| \ucalx{T}|} \bigg[m_F \gf(OPT) - \gf(\emptyset)\bigg]\\
    &\ge \left( 1 - \frac{\gamma_F}{q_{\max}} \right)^{q_{\max}} \bigg[m_F \gf(OPT) - \gf(\emptyset)\bigg]
\end{align}
Re-arranging this, we get the required result.
% \begin{align}
%     \gf( \ucalx{T}) &\le m_F \gf(OPT)-\left( 1 - \frac{\gamma_F}{q_{\max}} \right)^{q_{\max}} (m_F \gf(OPT)-\gf(\emptyset))
% \end{align}
\end{proof}

\section{Discussion on the assumptions}
\label{app:assump}
\textbf{(1)} Bounded difference between uncertainties across two feature subsets: Given a bucket $b\in [B]$, $|\Delta_i(\ucal{}) - \Delta_i(\vcal{})| \le \ee_{\Delta}$.  Note that since the uncertainties quantities are probabilities, the difference is always bounded between $0$ and $1$.
In our experiments, we found that $\Delta_i(U)$ is approximately around $0.8$ for most $U$ across different datasets. Hence, we set $\Delta_i(U) \approx 0.8$ directly. This automatically leads us $\ee_\Delta \approx 0$.
% The difference in general need not be very small. A small difference means that classifier confidence does not change too much with the generator. 

\textbf{(2)} Bounds on uncertainty and loss: 
$0< \Delta_{\min} \le |\Delta_i(\ucal{})| \le \Delta_{\max} $.
$ 0< \ell_{\min} \le |\ell(h_\theta(\xb_i),y_i)| \le \ell_{\max}$. 
In our setup, we found that $\ell_{\min} \in [0.003, 2.77]$ and  $\ell_{\max} \in [2.13, 8.99]$. As mentioned in item (1), we did not observe much variation of confidence values across datasets on the pre-trained models. This led to $\Delta_{\min}\approx 0.2$

\textbf{(3)} Lipschitzness: The loss function $\ell(h_\theta(\xb),y)$ is Lipschitz with respect to $\xb$. The activation functions within $h_{\theta}(\xb)$ 
are ReLU, which are Lipschitz. Moreover, $\ell$ is a  cross entropy loss, which is smooth. We tracked the gradient of different points, which revealed that $|\nabla_{\xb}\ell(h_\theta(\xb),y)|< L_x$ where $L_x \in [{3.00}, 4.15]$.

\textbf{(4)} Bounded difference between oracle and generated features: $\EE_{\xg_i}[||\xb_i-\xg_i \given \ocal{i}||] \le \ee_x$, for all $i$.  This difference need not be small--- the value of $\ee_x$ is small if the generator is near perfect.  The value of $\ee_x$ depends on how accurately the generator can mimic the oracle.  In our experiments, we found $\ee_x \in [3.80,14.19]$. Only in CIFAR100, we found that $\ee_x$ to be large with $\ee_x =  107.18$, since $\xb_i$ is high dimensional.

The computed approximation factor across all datasets came out to be less than a constant factor $\sim 10$ for our method. 
\newpage

\section{Additional Details about Experiments}
\label{app:setup}
This section provides further details about the datasets used, the implementation of our baselines, hyperparameter tuning and computing resources.

\subsection{Datasets}
\begin{table}[!htb]
    \centering
    \resizebox{0.8\textwidth}{!}{
\begin{tabular}{|l|c|c|c|c|c| }
 \hline
 Dataset & $|D|$ & \# of features = n &   $\EE[\ocal{}]$ &
 Budget $q_{\max}$ & optimal \# of buckets $B$\\ \hline
DP & 3,600 & 132 & 20 & 20-50 \% & 8 \\ \hline
MNIST & 60,000 & 784 & 40 & 5-17 \% & 8 \\ \hline
CIFAR100 & 60,000 & 256 & 20 & 20-50 \% & 4 \\ \hline
TinyImagenet & 110,000 & 256 & 20 & 20-50 \% & 4 \\ \hline
 \hline
\end{tabular}}
\caption{Dataset statistics}
\label{table:dataset}
\end{table}
\xhdr{Disease prediction (DP)} Disease prediction (DP) is a disease classification
dataset. Each feature indicates a potential disease symptom. Here the number of features $n=132$ and the classes $\Ycal$ indicate a set of $42$ diseases.  

\xhdr{MNIST}
The images of MNIST dataset are flattened into $n=784$ dimensional feature vectors, with $|\Ycal|=10$ classes. 

\xhdr{CIFAR100} We club together neighbouring pixel values into one feature, so that it has $n=256$ features. None of the baselines reported results on large datasets like CIFAR100 or Tinyimagenet. We found that these baselines are difficult to scale with larger dataset. Similar observations are also made by~\citet{gsm}.  The images in these datasets are not flattened as we use a CNN based architecture. 

\xhdr{Tinyimagenet} Here, $|\Ycal|= 200$. Similar to CIFAR100, we club together neighbouring pixel values into one feature, so that these two datasets have 256 features each. The images in these datasets are not flattened as we use a CNN based architecture. 

For MNIST, we take $q_{\max}$ in the range of $5-20 \%$, while for the other datasets it is $20-50 \%$. For MNIST, we observe that smaller number of features is enough for high accuracy.

Table~\ref{table:dataset} summarizes the dataset statistics.
% The reason for lower $q_{\max}$ for MNIST  
% the dataset has $784$ features, but \our manages to reach a high accuracy with only a few features.

\subsection{Details about the baselines}
We make use of the available implementations of the baselines: JAFA \footnote{\scriptsize \texttt{https://github.com/OpenXAIProject/Joint-AFA-Classification}}, EDDI\footnote{\scriptsize \texttt{https://github.com/Microsoft/EDDI}}, ACFlow\footnote{\scriptsize \texttt{https://github.com/lupalab/ACFlow-DFA}}, GSM\footnote{\scriptsize \texttt{https://github.com/lupalab/GSMRL/tree/GSMRL}}, CwCF\footnote{\scriptsize \texttt{https://github.com/jaromiru/cwcf}}. We express gratitude to the authors of DiFA for providing us with code for its implementation. In the sequential setup, they query features one by one, where they query feature $x[u_1]$ from the oracle at time $t$, observe its value and then use all the features observed until time $t$ ($\xb[\ocal{} \cup u_1\cup u_2...\cup u_t]$) to query the next feature at time $t+1$.

We also modify these baseline to function in a batch setting: \textbf{(1)} EDDI and ACFlow make use of a greedy algorithm over the features. We thus acquire the top-$q_{\max}$ features \emph{all-together} using the greedy algorithm starting from $\xb[\ocal{}]$, without querying them one-after-another.  \textbf{(2)} JAFA, GSM, CwCF and DiFA have a RL policy to pick features. The policy has its action space as the set of features and the termination action. We make use of the top-$q_{\max}$ features from the policy when the MDP is in the $\obs$ state.
%

% The off-the-shelf implementation of the baselines shows poor performance especially on the larger dataset. \ad{VA: show results for one larger dataset, TI maybe?}
Some of the baselines perform poorly on the larger datasets as they are not designed to scale for large dataset. Therefore, we deployed the mixture of experts model on the baselines. Specifically, we performed the same  partitioning on the dataset and then trained the mixture of experts using the features obtained from underlying baseline, during training.  

% on the same partitioning of the data, whereas we trained 

% Hence, we take the features that these baselines choose and deploy a mixture of experts model on them. We report the best among the off the shelf results and the hybrid results.

\subsection{Additional details about the generator $p_{\phi}$}
\xhdr{DP and MNIST} The generator consists of an encoder $\text{Enc}(\bm{\eta}  \given \xb)$ and a decoder $\text{Dec}(\xb \given \bm{\eta})$, which are trained like a $\beta$-VAE during the initial stage before bucketing (Algorithm~\ref{alg:training}, function \textsc{GreedyForU}, line 4). We set the VAE regularization parameter $\beta$ to be equal to $\frac{\sqrt{2}}{100}n$, where $n$ is the number of features. Specifically, we train:
\begin{align}
\sum_{i\in D}\EE_{\bm{\eta}_i\sim\text{Enc}(\bm{\eta}_i \given \xb_i[\ocal{i}])} \log \text{Dec} (\xb[\ocal{i}]\given \bm{\eta}_i) +  \beta KL(\text{Enc}(\bullet\given\xb_i[\ocal{}]) || \text{Prior}(\bullet))
\end{align}
We choose $\text{Prior}(\bm{\eta}_i) = \text{Normal} (0,\II)$.

The encoder $\text{Enc}(\bm{\eta}_i \given \xb_i[\ocal{i}])$ for DP and MNIST datasets is a transformer, with $7$ self-attention heads and $5$ self-attention layers. We train a position embedding $W \in \mathbb{R}^{n \times d_z}$, such that for each position $\text{pos}\in \ocal{i}$, the corresponding position embedding is given by $W^T \II_{\text{pos}}$, where $\II_{\text{pos}}$ is the one-hot encoding for position $\text{pos}$. The length of the position embedding $d_z$ for each position is set to be 20 and 69 respectively for DP and MNIST. The position embedding undergoes position wise concatenation with the feature vector, before it is fed to the transformer encoder. 
% Specifically, given $\xb_i[\obs_i]$, the encoder produces a set embedding $\zb_{i}$ of shape $|\obs_i| \times d_{\zb}$ using pretrained position embeddings, where we set $d_{\zb} = 20$ for \dpr\ and $d_{\zb} = 69$ for \mnist. 
% The transformer then takes the position wise concatenation of $\xb$ and $\zb$ and produce $\bm{\eta}_i= \text{Concat}\left([\xb_i[\text{pos}],\zb[\text{pos}]]\right)_{\text{pos} \in \obs_i}$ as output. 
Thus, we have the following latent code generation process:
\begin{align}
&\zb_i = W^T[\II_{\text{pos}_1} \II_{\text{pos}_2} \cdots \II_{\text{pos}_{|\ocal{i}|}}]_{pos \in \ocal{i}}\\
&\bm{\eta}_i=  \text{Transformer}(\text{Concat}\left([\xb_i[\text{pos}],\zb_i[\text{pos}]]\right)_{\text{pos} \in \obs_i})
\end{align}

The decoder $\text{Dec}$ is a cascaded network of linear, ReLU and linear layers with a hidden layer size of $32$. 

\xhdr{CIFAR100, TinyImagenet} For CIFAR100 and TinyImagenet datasets, we make use of Resnet-152 encoder and a DCGAN decoder. Any feature $\xb_i[\ocal{i}]$ is fed into the Resnet-152 encoder to obtain the embedding $\bm{\eta}_i$. This is further fed into the DCGAN decoder to generate $\xb_i$. In all cases, observed features are encoded by masking the input image. 

% \textbf{(6)} Cascaded network of linear, ReLU and linear layers, are used as a classifier for DP and MNIST, which receive the encoder's output as input. \textbf{(7)} CIFAR100 makes use of WideResnet, while TinyImagenet uses EfficientNet as classifier.

\subsection{Hyperparameter selection}
The number of buckets in the random hyperplane based RHclustering is chosen by experimenting with different sizes. We use 8,8,4 and 4 buckets for DP, MNIST, CIFAR100 and TI datasets respectively.
In case of DP and MNIST, we make use of Adam optimizer with a learning rate of $10^{-3}$. The other two datasets use SGD optimizer, with a learning rate schedule. During inference, we set the threshold hyperparameter $\tau$ (Inference algorithm, Algorithm~\ref{alg:inference}) such that 
top $10\%$ instances are chosen to make use of $\xg[\vcal{}]$.

We make use an embedding dimension of $100$ for the generator of EDDI. JAFA makes use of $10000$ pretraining steps and $5\times 10^{5}$ episodes of training. For ACFlow and GSM, we make use of $256$ dimensional layers for the affine transformation as well as the coupling. We train the flow models for $500$ epochs. We evaluate CwCF with the trade-off parameter $\lambda$ set to 1. DiFA is executed with $1000$ pretrain iterations and $4000$ training iterations, and gradient norms set to $10$ or $100$ for each of the datasets.

\xhdr{Software and Hardware details}
We implement our method using Python 3.8.10 and PyTorch 1.13.1. Model training for \our\ and baselines was performed on two servers: (1) 16-core Intel(R)693
Xeon(R) Gold 6226R CPU@2.90GHz with 115 GB RAM,  containing Nvidia RTX A6000-48 GB; and,
(2) NVIDIA DGX servers containing eight A100-80 GB GPUs.

\xhdr{License Details}
EDDI is available under Microsoft Research License. The DP, TinyImagenet and CIFAR100 datasets and CwCF implementation are obtained under MIT license, while MNIST is obtained under GNU license.

\newpage
\section{Additional experiments}
\label{app:exp}
\subsection{Analysis of standard error}
\begin{figure}[h]
    \centering
    \includegraphics[width=0.5\linewidth]{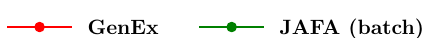}\\
\subfloat[DP]{\includegraphics[width=.24\textwidth]{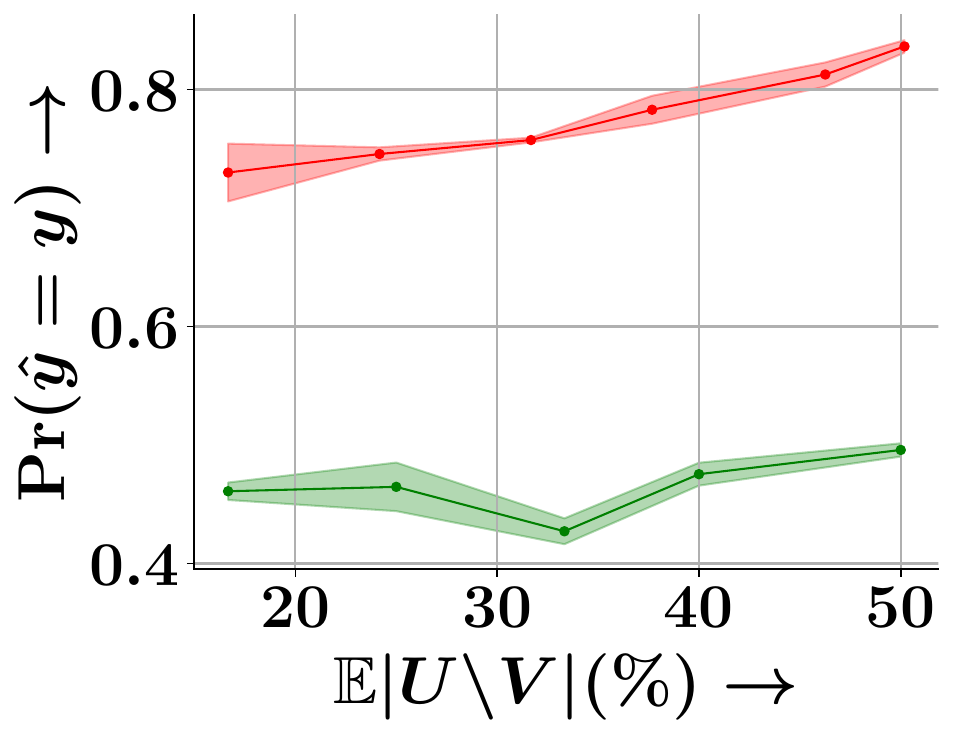}} \hspace{0.5mm}
\subfloat[MNIST]{\includegraphics[width=.23\textwidth]{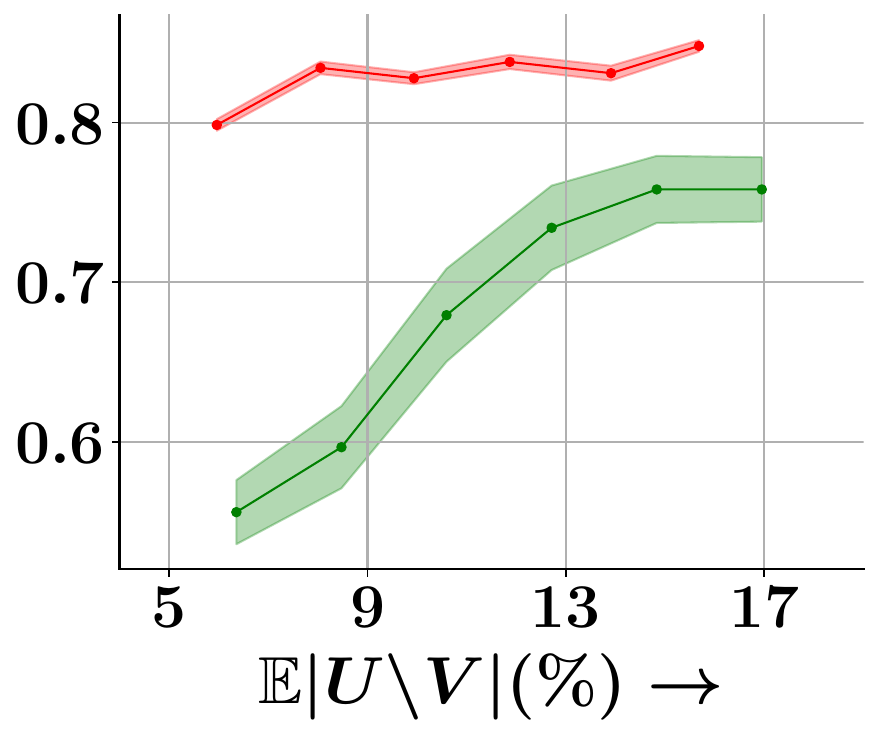}}\hspace{0.5mm}
\subfloat[CIFAR100]{\includegraphics[width=.23\textwidth]{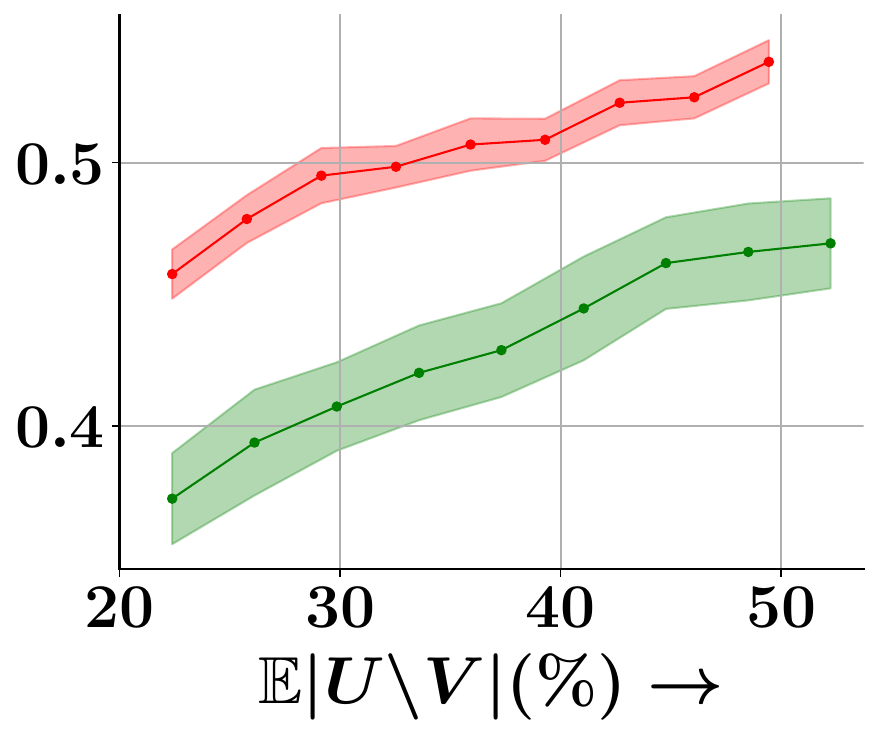}}\hspace{0.5mm}
\subfloat[TinyImageNet]{\includegraphics[width=.23\textwidth]{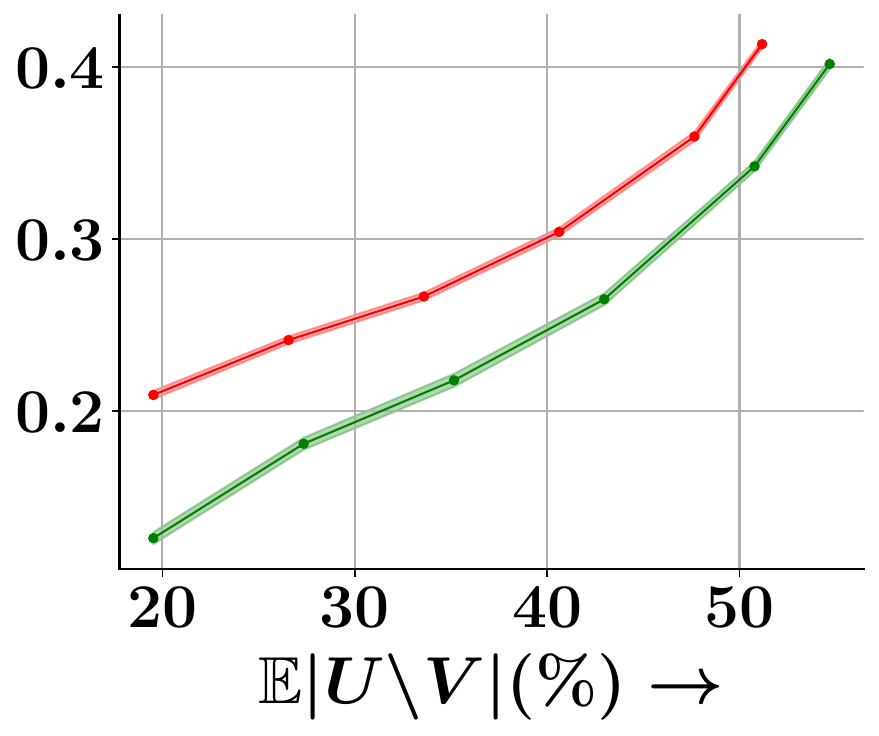}}
\vspace{-1mm}
\caption{Plot of mean accuracy and standard error for \our\ and JAFA (batch), evaluated using $20$ monte-carlo samples of the accuracy.}\vspace{-1mm}
\label{fig:error_bars}
\end{figure}

Here, we plot the standard error and analyze the statistical significance of our results. In Figure \ref{fig:error_bars} we plot the mean and standard error in accuracy calculated on $20$ fold cross validation for \our\ and the baseline that performs closest to our method, JAFA (batch). 
We also perform Welch t-test to look into how much significance our performance improvements have. 
In Tables \ref{table:p_values_dp} and \ref{table:p_values_mnist}, we also present the corresponding p-values.  We observe: \textbf{(1)} \our\ provides a statistically significant improvement in accuracy, as the p-values are all $<0.01$, while in majority of the cases p-values are less than $10^{-3}$.  \textbf{(2)} \our\ has less standard error in accuracy than JAFA (batch), showing that our method performs consistently.
\begin{table}[!htb]
    \centering
\begin{tabular}{ |c|c|c|c|c|c| } 
\hline 
$\EE|\ucal{} \cp \vcal{}|$ & $\approx 20 \%$ & $\approx 30 \%$ & $\approx 40 \%$ & $\approx 50 \%$ \\
 \hline
 DP & $10^{-23}$ & $10^{-18}$ & $10^{-11}$ & $10^{-19}$ \\ 
 \hline
  CIFAR100 & $4.6\times 10^{-5}$ & $2.4\times 10^{-5}$ & $ 0.0021$ & $ 0.0024$ \\ 
 \hline
 TinyImaganet & $10^{-15}$ & $10^{-13}$ & $10^{-10}$ & $3\times10^{-6}$ \\ 
 \hline
\end{tabular}
\caption{p-values for the t-test to measure the statistical significance of the performance gain by our method as compared to the nearest baseline (JAFA (batch)) for DP, CIFAR100 and TinyImagenet.}
\label{table:p_values_dp}

\begin{tabular}{ |c|c|c|c|c|c|c| } 
\hline 
$\EE|\ucal{} \cp \vcal{}|$ & $\approx 7 \%$ & $\approx 9 \%$ & $\approx 11 \%$ & $\approx 13 \%$ & $\approx 15 \%$ & $\approx 17 \%$\\
 \hline
 MNIST & $10^{-10}$ & $10^{-8}$ & $3.2 \times 10^{-5}$ & $0.0005$ & $0.0013$ & $0.0001$ \\ 
 \hline
\end{tabular}
\caption{p-values for the t-test to measure the statistical significance of the performance gain by our method as compared to the nearest baseline (JAFA (batch)) for MNIST.}
\label{table:p_values_mnist}
\end{table}

\subsection{Performance variation across number of clusters}
\begin{figure}[h]
    \centering
    \includegraphics[width=0.5\linewidth]{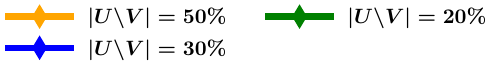}\\[-1ex]
 \subfloat[DP]{\includegraphics[width=.25\textwidth]{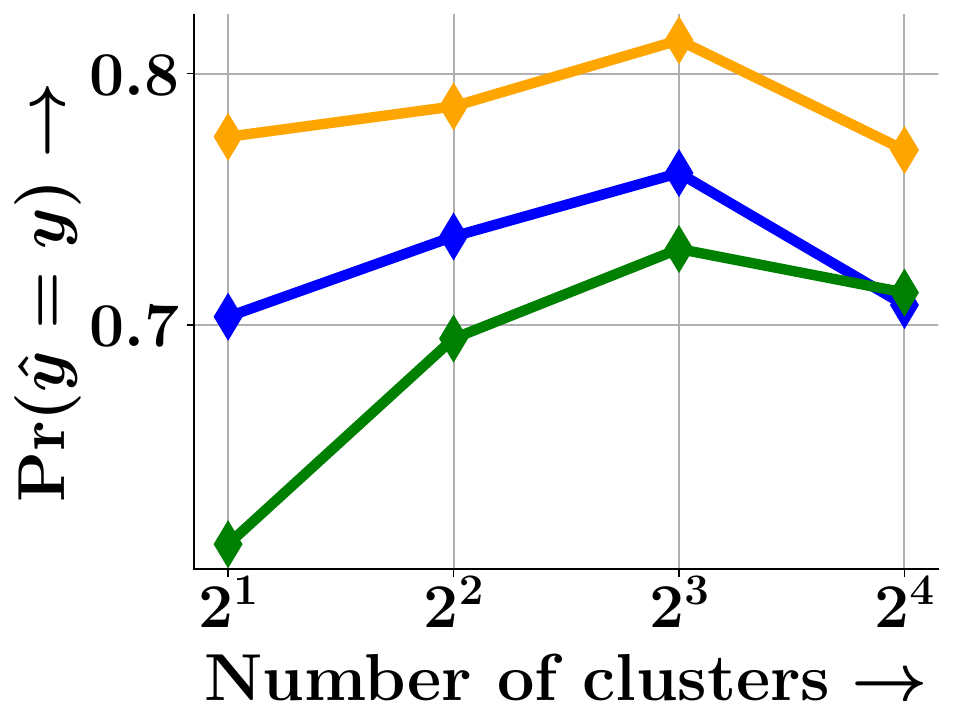}}  
\subfloat[CIFAR100]{\includegraphics[width=.25\textwidth]{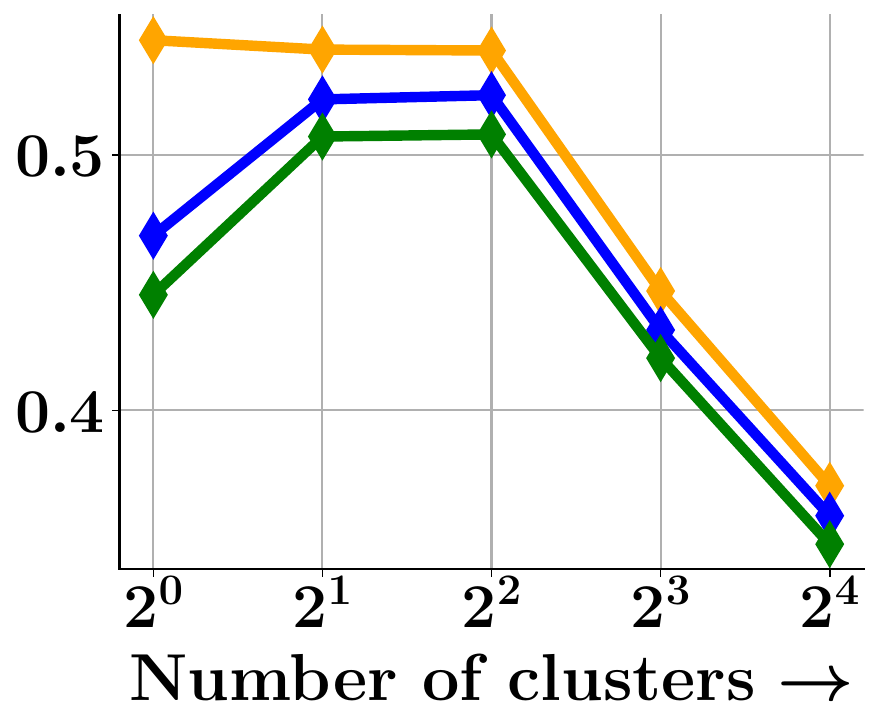}}
\vspace{-1mm}
        \caption{Change in classification accuracy with variation of number of buckets $|B|$. Shows a peaking behaviour.}
        \label{fig:clusters}
\end{figure}
We consider the effect of varying the number of clusters in RH clustering on the prediction accuracy 
against varying $|\ucal{} \cp \vcal{}|$ in Figure \ref{fig:clusters}. It is observed to first increase with the number of buckets, followed by a decrease. \textbf{(1)} The increase can be explained by the reduction in heterogeneity in each buckets as the number of buckets rise. \textbf{(2)}  However, the accuracy peaks and then falls off, as with decrease in the training set size in each bucket, the mixture of experts model becomes less robust to noise. Hence, its performance decreases on the test set with the noisy features produced by the generator.

\subsection{Sequential variants of baselines}

\begin{figure*}[!h]
    \centering
    \includegraphics[width=0.8\linewidth]{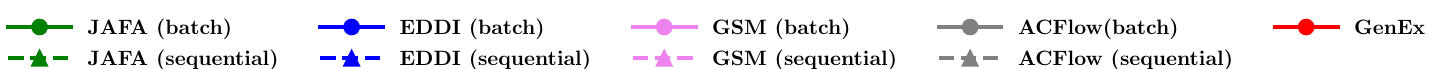}\\
\subfloat[DP]{\includegraphics[width=.18\textwidth]{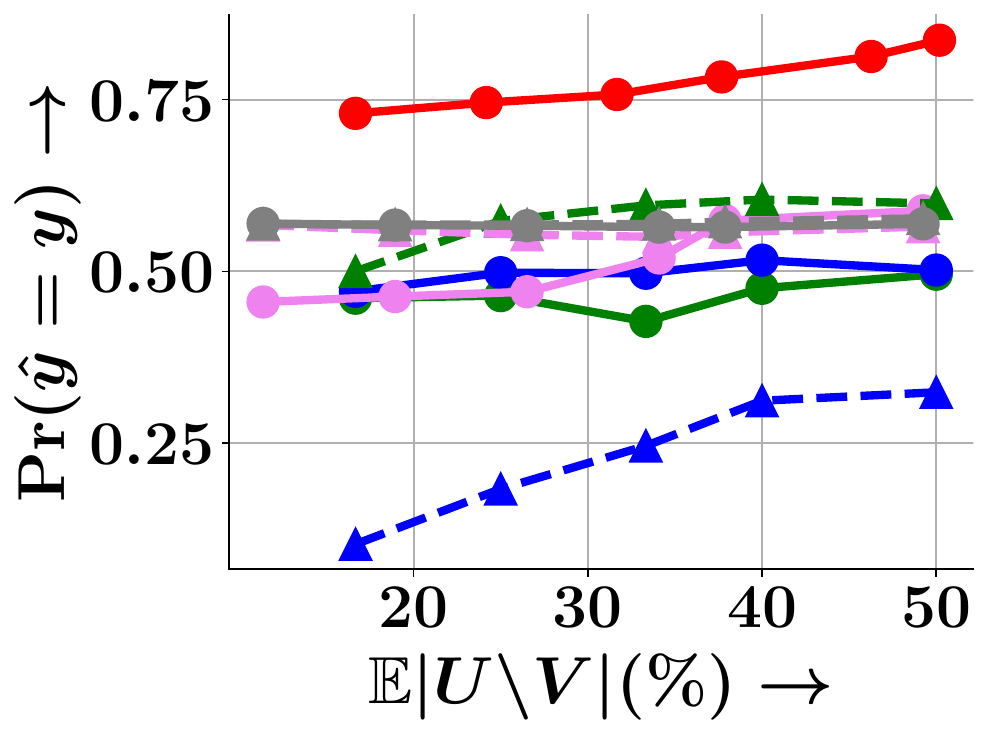}} \hspace{0.5mm}
\subfloat[MNIST]{\includegraphics[width=.17\textwidth]{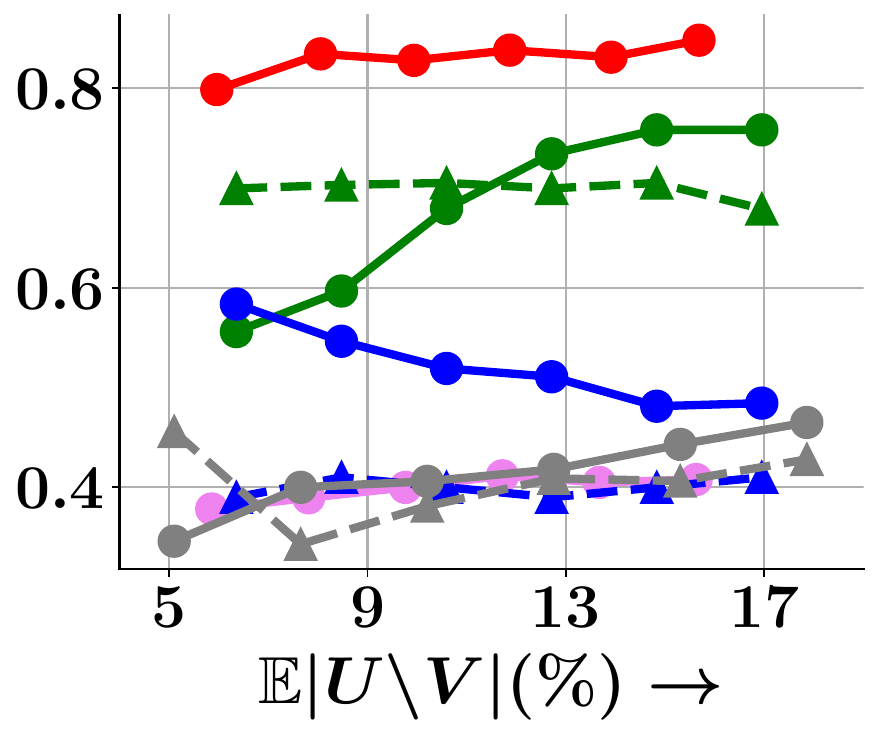}}\hspace{0.5mm}
\subfloat[CIFAR100]{\includegraphics[width=.17\textwidth]{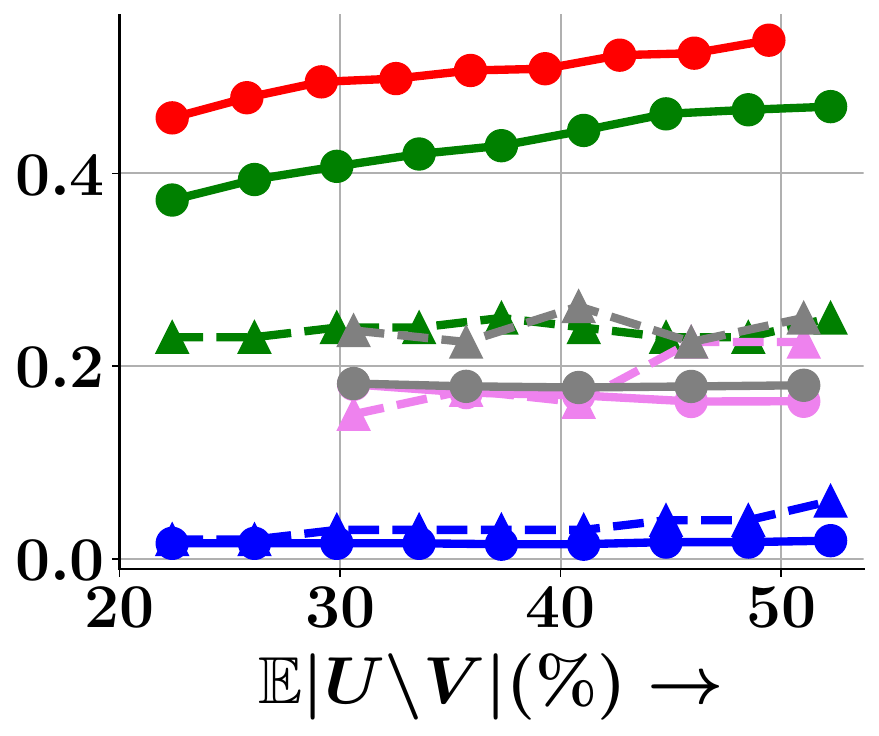}}\hspace{0.5mm}
\subfloat[TinyImageNet]{\includegraphics[width=.17\textwidth]{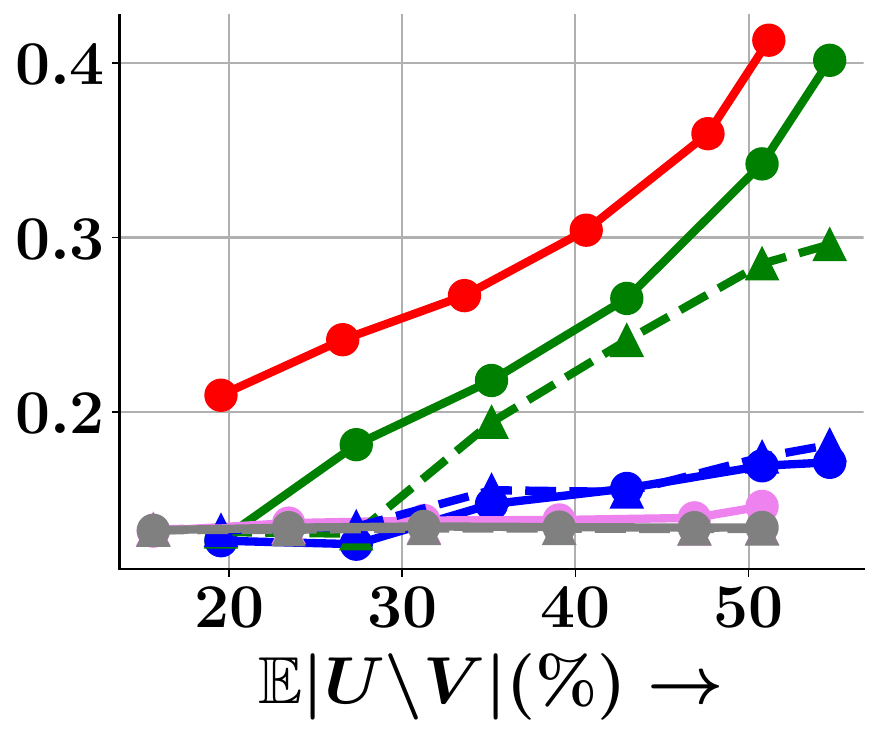}}
\vspace{-1mm}
\caption{Comparison  of \our\ against batch and sequential variants of baselines, \ie, JAFA \cite{jafa}, EDDI \cite{eddi}, ACFlow \cite{acflow} and GSM \cite{gsm}, in terms of the classification accuracy varying over the average number of oracle queries $\EE|\ucal{} \cp \vcal{}|$, for all four datasets.}\vspace{-1mm}
\label{fig:main_full}
\end{figure*}

We also compare \our\  the sequential variants of the baselines. The baselines can now acquire features one-by-one, observing the feature value before querying for the next feature. We illustrate the results for EDDI, JAFA, ACFlow and GSM in Figure \ref{fig:main_full}. The performance of the sequential variants of the baselines are sometimes worse than their batch variants. 

It is natural to expect that the sequential variant of an algorithm should outperform the batch variant, since more information is available at every stage of feature selection in a sequential algorithm. However, Figure~\ref{fig:main_full} demonstrates that batching does not perform too poorly. Infact, GENEX which employs batching outperforms the baselines significantly.
These baselines are not designed to scale to a very large number of features (as asserted in the classification experiments of~\citet{gsm}), and hence their accuracy stagnates after a few features. Evidence of this can also be found in GSM~\citet[Figs. 6,7]{gsm}, which deals with a similar set of baselines. The stagnation of JAFA with an increasing number of papers can also be seen in figure 3~\cite{jafa}.

We observe that sequential ACFlow even drops  when $\EE[U\cp V]$ goes from 5 to 8.  We posit that it arises due to ACFlow being used in a batch setting during inference time, while its training algorithm makes use of a sequential learning process. As a result it is unable to identify a good set of features in the batch setting and the accuracy drops.

% It was surprising to us, since the sequential variant can use more information than the batch variant, by utilizing the oracle value of a queried feature as feedback to query next.  The batch variants of the baselines offer more diversity in the acquired features, since the features acquired in the sequential variants are correlated with each other.

\subsection{Additional ablation study: Use of MOE instead of one single monolithic model improves the performance of baselines}
\begin{figure}[h]
    \centering
      \includegraphics[width=0.5\linewidth]{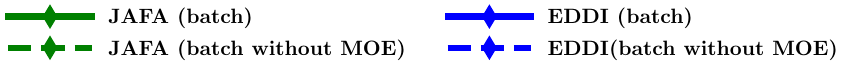}\\[-1ex]
 \subfloat[DP]{\includegraphics[width=.25\textwidth]{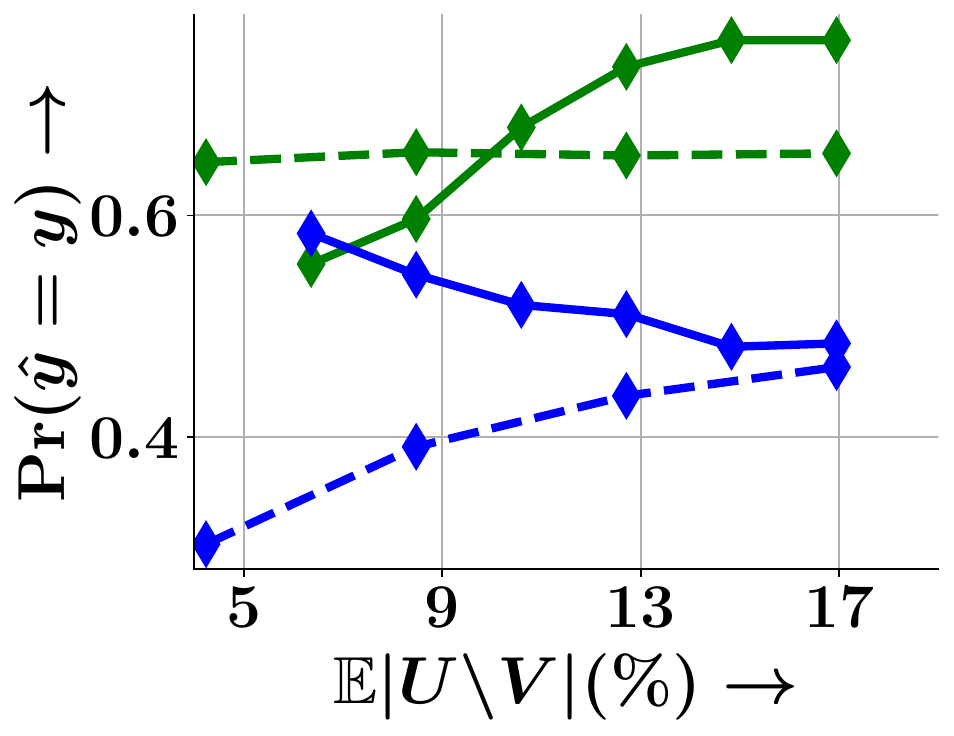}}  
\subfloat[CIFAR100]{\includegraphics[width=.25\textwidth]{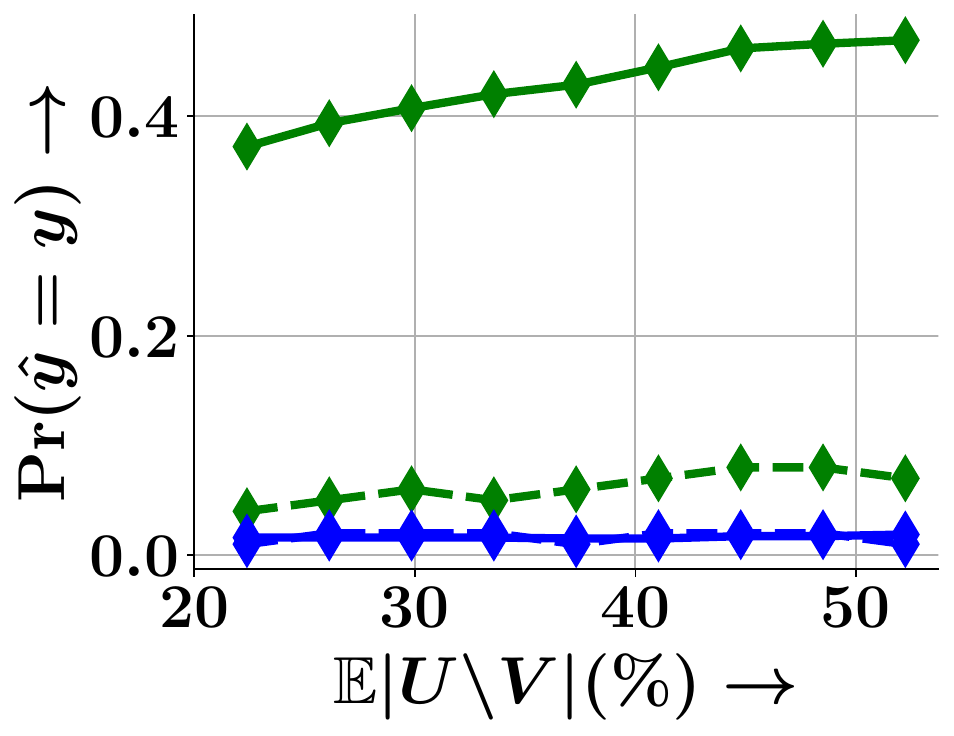}}
\vspace{-1mm}
\caption{Comparison of accuracies for JAFA (batch) and EDDI (batch) against hybrid variants.}
    \label{fig:hybrid}
\end{figure}

In some cases, the baselines fail to scale to the large number of features that we provide to them. As a result, their classifiers are not able to train properly. To give the baselines a fair stage to compete, we deploy a mixture of experts model on the feature subsets that the baselines choose. We then report the best among the off the shelf baseline and the mixture of experts variant. This is done for both sequential and batch variants of the baselines. We illustrate this for the MNIST and CIFAR100 datasets in Figure \ref{fig:hybrid}.

% \begin{wrapfigure}[13]{r}[-0pt]{0.3\columnwidth}
% \vspace{-3mm}
%     \centering

% \end{wrapfigure}

\subsection{Classifier behaviour on generated data}
\begin{figure}[h]
    \centering
 \subfloat{\includegraphics[width=0.22\textwidth]{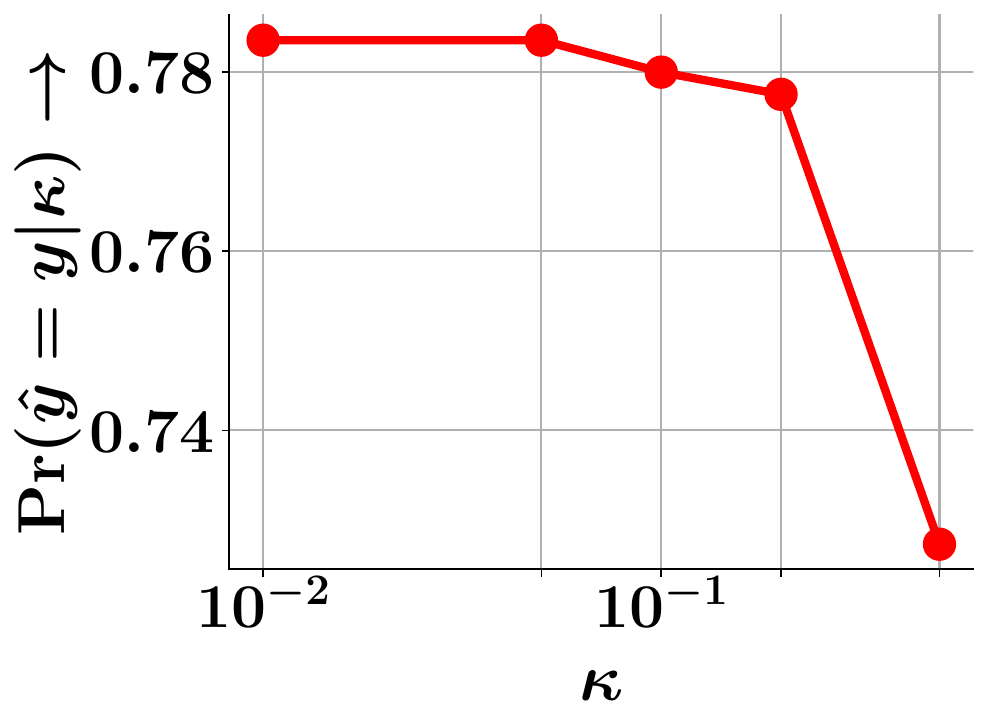}}   
\subfloat{\includegraphics[width=0.22\textwidth]{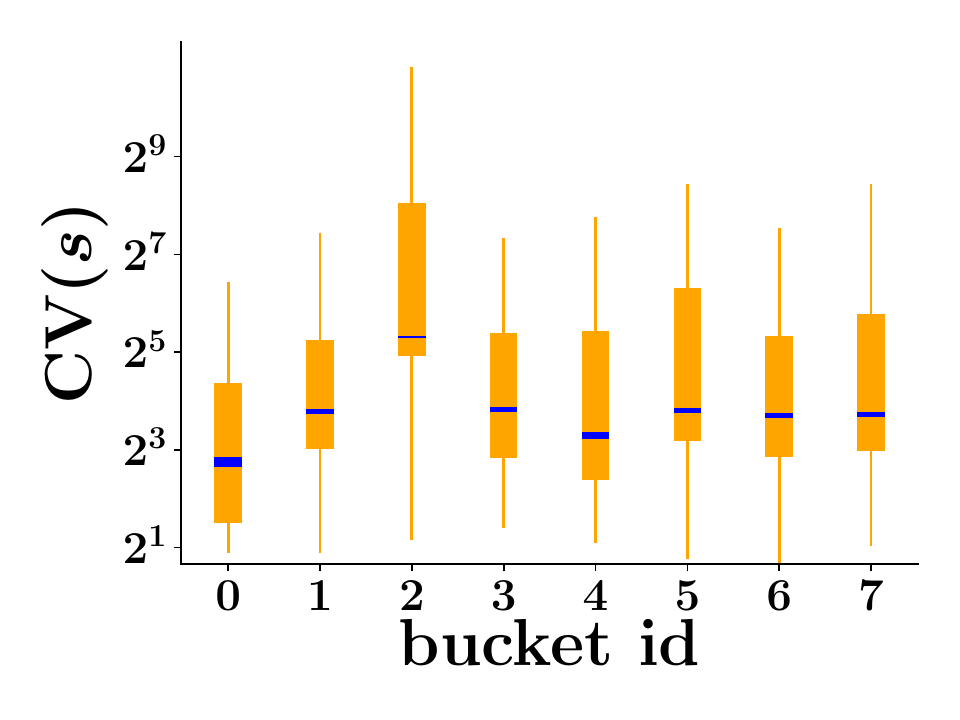}} \vspace{-3mm}
        \caption{\small Analysis of $\EE[|x[s]-x'[s]|]$ }
    \label{fig:lm}
\end{figure}
\textbf{Can the set of accurate features generated by $p_{\phi}$ be used for prediction?}
We consider the behaviour of the classifier on generated data when the generator produces accurate features.
% We demonstrate that the closeness of generated features to the gold data isn't always a suitable metric to choose features.
Here we attempt to use the difference of oracle and generated features $\ee_x= ||\xb-\xg||$. We plot the variation of $\Pr(y=\hat{y} \given \kappa)$ against $\kappa$, where we use generated features on points which constitute the bottom $\kappa$ fraction of the population in terms of $\ee_x$. The remaining points make use of oracle features. Figure~\ref{fig:lm} (left) shows that for instances with low $\kappa$, the accuracy is consistently high. 

This immediately triggers the possibility of designing a predictor of $\xb - \xg$, so that we can use generator for only those points where this difference small. But is  $\ee_s = x[s] - x'[s]$ predictable?  To investigate this, we compute $\ee_s$ for all $i\in D_b$ for each bucket $b$. Then, we compute the coefficient of variation CV$(s) = \text{std}_s/\text{mean}_s$ where, 
$\text{std}_s, \text{mean}_s$ are the population standard deviation and mean of $\ee_s$ across different
instances in $ D_b$. We plot CV$(s)$  for all buckets $b$ in Figure \ref{fig:lm} (right). It shows that the CV$(s)$ is quite large making $\ee_s$ prediction infeasible.

\subsection{Alternative inference algorithm}
We consider an alternative to using the classifier's confidence in the inference algorithm. We train a supervised classifier $\tau_{\psi}$. This takes $\xb[\ocal{} \cup \ucal{}\cp\vcal{}]\cup \xg[\vcal{}]$   as input and  predicts if the trained classifier $  h_{\hat{\theta}_b}$ gives a correct prediction on $\xb_i[\ocal{i} \cup \ucal{i}\cp\vcal{i}]\cup \xg_i[\vcal{i}]$
Specifically, $\tau_{\psi}$ makes the following prediction:
\begin{align}
\tau_{\psi}(\xb[\ocal{} \cup \ucal{}\cp\vcal{}]\cup \xg[\vcal{}]) = \mathbf{1}\left[\max_{y'}   h_{\hat{\theta}_b}(\xb[\ocal{} \cup \ucal{}\cp\vcal{}]\cup \xg[\vcal{}])[y']=y\right] 
\end{align}
Hence, the classifier $\tau_{\psi}$ is trained 
on the pairs $(\xb_i[\ocal{i} \cup \ucal{i}\cp\vcal{i}]\cup \xg_i[\vcal{i}], r_i)_{i\in D}$, where $r_i = 1$ if the trained classifier gives correct prediction on $\xb_i[\ocal{i} \cup \ucal{i}\cp\vcal{i}]\cup \xg_i[\vcal{i}]$ and $r_i =0$, otherwise. Once trained,  we use $\xg[V]$ instead of oracle features $\xb[V]$, only for those instances, where $\tau_{\hat{\psi}}(\xb[\ocal{} \cup \ucal{}\cp\vcal{}]\cup \xg[\vcal{}]) =1$. The architecture of $\tau_{\psi}$ is taken to be the same as that of the classifier. Hence, for DP and MNIST datasets, we take $\tau_{\psi} (\bullet) =  \text{Sigmoid}(\text{Linear}(\text{ReLU} (\text{Linear}(\text{Encoder}(\bullet))))$ as the underlying architecture for $\tau_{\psi}$. And for CIFAR100 and TI, the architectures are WideResnet and EfficientNet respectively.

\begin{figure}[h]
    \centering
      \includegraphics[width=0.5\textwidth]{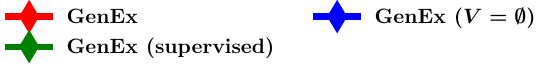}\\[-0.5ex]
\includegraphics[width=.3\textwidth]{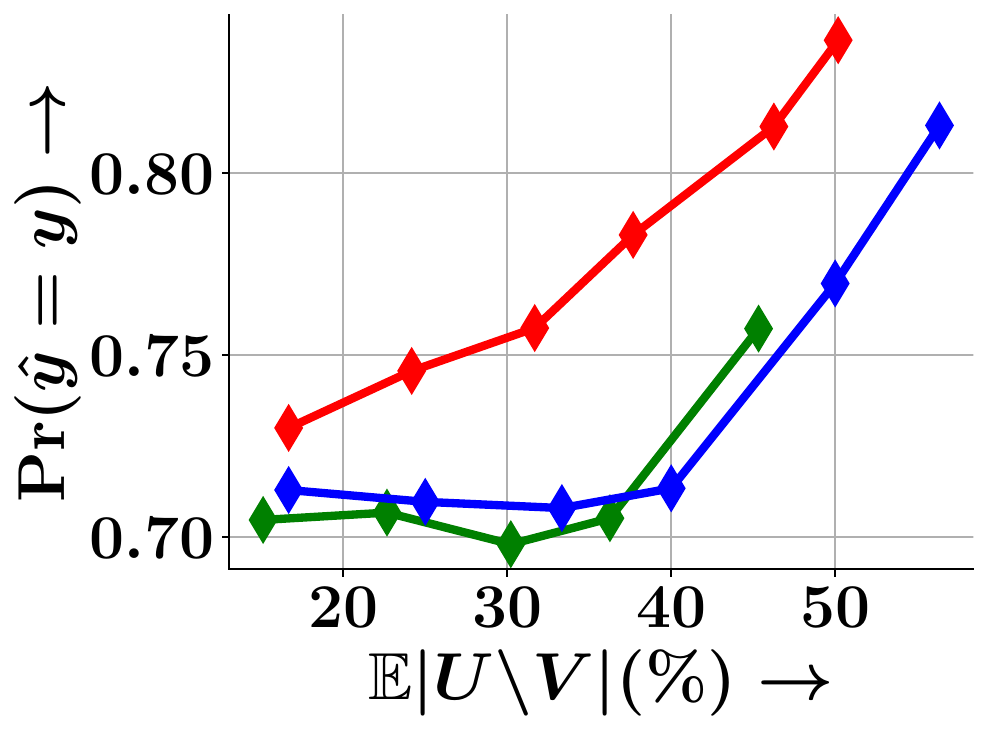}
\caption{\small Comparison of accuracy of \our, \our(supervised) and \our($V=\emptyset$)}
\label{fig:choice_dp}
\end{figure}

 Accuracy results of \our, \our(supervised) and \our($V=\emptyset$) are in Figure \ref{fig:choice_dp}. The following can be observed: \textbf{(1)} \our\ outperforms the \our(supervised) variant. This is because the supervised classifier has only $\approx 70 \%$ accuracy at identifying data points where the classifier would do well on generated features. \textbf{(2)} \our(supervised) does better than \our($V=\emptyset$) at higher budget as it is able to save cost more effectively.

\newpage

\subsection{Efficiency analysis}
We compare the training time per epoch and test times of \our\ against the baselines in Tables \ref{table:train_time} and \ref{table:test_time} for the smallest and the largest dataset. 
% In each case, we performed a distributed training and test for all the methods  across different buckets. 
In each case, the numbers are reported without any parallelization. 
We see that \our\ performs competitively with most baselines. \textbf{(1)} \our's speed is due to its simple inference strategy of identifying a data point's cluster, and then utilizing the $\ucal{},\vcal{}$ for that bucket, which does not need any significant computation. \textbf{(2)} EDDI's better training time can be attributed to the fact that their generator model is trained with random subsets of features, and no greedy method is employed during training. However this is offset by \our's superior performance during inference as EDDI performs greedy acquisition without applying any trick to speedup.
\begin{table}[!htb]
    \centering
    \begin{tabular}{ |c|c|c|c|c|c|c|c| } 
 \hline
 Dataset & \our & EDDI & JAFA & GSM & ACFlow & CwCF & DiFA \\ 
 \hline
 DP & 32 s & 10 s & 140 s & 65 s & 50 s & 30 s & 300 s\\ 
 \hline
 TI & 2.5 hrs & 2.5 hrs & 3 hrs & 3 hrs & 2.75 hrs & 2.5 hrs & 3 hrs \\ 
 \hline
\end{tabular}
\caption{Training time per epoch. We train each method without any parallelization with a fixed batch size (1024) across all the methods. For JAFA, GSM, CwCF and DiFA, we consider an epoch as collection of RL episodes which cover the whole dataset exactly once.}
        \label{table:train_time}

\begin{tabular}{ |c|c|c|c|c|c|c|c| } 
 \hline
 Dataset & \our & EDDI & JAFA & GSM & ACFlow & CwCF & DiFA  \\ 
 \hline
 DP & 62 s & 15 min & 100 s & 60 s & 240 s & 60 s & 80 s \\ 
 \hline
 TI & 20 min & 2 hrs & 13 min & 20 min & 16 min & 15 min & 20 min \\ 
 \hline
\end{tabular}
\caption{Test time for $|\ucal{}\cp \vcal{}| \approx 50\%$. We test serially with a fixed batch size (1024). }
        \label{table:test_time}
\end{table}

All numbers are reported on experiments on 16-core Intel(R)693
Xeon(R) Gold 6226R CPU@2.90GHz with 115 GB RAM and one A6000 NVIDIA GPU with 48GB RAM.  

\subsection{Performance of EDDI in low dimensional datasets}

{We perform experiments with the gas detection dataset from kaggle. It is a classification task which uses 48 features (readings from sensors) to classify a gas sample into 6 classes.} For $|U/V|=35$. Following is the result.

\begin{table}[!ht]
    \centering
    \begin{tabular}{|l|l|l|}
    \hline
        GENEX & JAFA & EDDI \\ \hline
        0.94 & 0.62 & 0.41 \\ \hline
    \end{tabular}
\end{table}
 
\subsection{Position of observed feature in the image vs accuracy}

 In real life scenarios like medical diagnosis, the observed features have the semantics of being easily observed symptoms. However, here we do not assign any semantics to the features. We choose random subsets as $\Ocal$ to demonstrate the effectiveness of the GENEX algorithm over a wide range of subsets. We have a sufficiently large number of random subsets $\Ocal$, 20, 10, 5, 5 for the disease prediction, MNIST, CIFAR100 and TinyImagenet datasets. This is further validated using the error bars and p-values. 

To look at the role of observed features in the image datasets, we contrast the cases where the observed features lie at the center of the image v/s the observed features are towards the edge of the image in the CIFAR100 dataset

\begin{table}[!ht]
    \centering
    \begin{tabular}{|l|l|l|l|l|}
    \hline
        Position of $O$ &  $|U \backslash V|$ = 20 \% &$|U\cp V|$ = 30\% &$|U\cp V|$ = 40\% &$|U\cp V|$ = 50\% \\ \hline
        Center & 0.29 & 0.35 & 0.39 & 0.43 \\ \hline
        Edge & 0.37 & 0.43 & 0.46 & 0.48 \\ \hline
    \end{tabular}
    \caption{Accuracy for different positions of $\Ocal$ and different number of oracle queries}
\end{table}

This indicates that if the initial set of features is itself informative (like the center of image), we do not gain much accuracy by querying. If instead the initial set is not informative, we can make big gains in accuracy via querying of informative features.  

\newpage
\subsection{Comparison with other baselines}

In the following, we use~\citet{xiao2022group} and an unsupervised method which is facility location. Facility location is an unsupervised method that maximizes diversity among chosen features. Following are the results for the disease prediction dataset.

\begin{table}[h]
    \centering
    \begin{tabular}{l|l|l|l}
    \hline
Model&	$|U\cp V| = 20$\% &	$U\cp V=30\%$ & 	$|U\cp V=40\%$ \\ \hline
GENEX&	0.74	&0.76	&0.78 \\ \hline
Facility location	&0.45&	0.56&	0.64  \\ \hline
\citet{xiao2022group} &	0.69&	0.71&	0.71 \\ \hline
% Liyanage et al (Paper 3)	&0.02&	0.02	&0.02 \\ \hline
\end{tabular}
    \caption{Accuracy for variouy methods in DP dataset}
\end{table}
GENEX outperforms other methods.

% \begin{center}
% \Large{Author response to Neurips review}
% \end{center}
% \normalsize

% \newpage
% \bibliography{afa,refs}
% \bibliographystyle{plainnat}

\end{document}